%% file: PINeRF.tex
\documentclass[sigconf]{acmart}

\acmSubmissionID{880} 
\setcitestyle{nosort,square} 

\usepackage{xcolor, colortbl}
\usepackage{tikz}
\usetikzlibrary{shapes.geometric, arrows}

\usepackage{cuted}
\usepackage{tabularx}

\copyrightyear{2024}
\acmYear{2024}
\setcopyright{acmlicensed} 
\acmConference[SIGGRAPH Conference Papers
'24]{Special Interest Group on Computer Graphics and Interactive Techniques
Conference Conference Papers '24}{July 27-August 1, 2024}{Denver, CO, USA}
\acmBooktitle{Special Interest Group on Computer Graphics and Interactive
Techniques Conference Conference Papers '24 (SIGGRAPH Conference Papers
'24), July 27-August 1, 2024, Denver, CO, USA}
\acmDOI{10.1145/3641519.3657483}
\acmISBN{979-8-4007-0525-0/24/07}

\usepackage{hyperref}
\usepackage{url}
\usepackage{comment} 
\usepackage{multirow}

\definecolor{purple}{rgb}{0.5,0.0,0.97}
\definecolor{cyan}{rgb}{0.0,0.6,0.8}
\definecolor{green}{rgb}{0.3,0.56,0.0}

\newcommand{\myreffig}[1]{Fig.~\ref{#1}}

\newcommand{\resizeEq}[3]{
\begin{equation}
\resizebox{#3}{!}{%
$\begin{array}{@{}cc}
\begin{split}
#1
\end{split}
\end{array}$}  \label{#2}
\end{equation}
}
\newcommand{\vect}[1]{\boldsymbol{#1}}

\begin{document}
\title{Physics-Informed Learning of Characteristic Trajectories for Smoke Reconstruction}

\author{Yiming Wang}
\email{wangyim@ethz.ch}
\affiliation{
\institution{ETH Zurich}
\country{Switzerland}
}
\author{Siyu Tang}
\email{siyu.tang@inf.ethz.ch}
\affiliation{
\institution{ETH Zurich}
\country{Switzerland}
}
\author{Mengyu Chu}
\email{mchu@pku.edu.cn}
\affiliation{
\institution{Peking University, SKL of General AI}
\country{China}
}
\renewcommand\shortauthors{Wang, Y.; Tang, S.; Chu, M.}

\begin{abstract}
We delve into the physics-informed neural reconstruction of smoke and obstacles through sparse-view RGB videos, tackling challenges arising from limited observation of complex dynamics. 
Existing physics-informed neural networks often emphasize short-term physics constraints, leaving the proper preservation of long-term conservation less explored. 
We introduce Neural Characteristic Trajectory Fields, a novel representation utilizing Eulerian neural fields to implicitly model Lagrangian fluid trajectories. 
This topology-free, auto-differentiable representation facilitates efficient flow map calculations between arbitrary frames as well as efficient velocity extraction via auto-differentiation.
Consequently, it enables end-to-end supervision covering long-term conservation and short-term physics priors.
Building on the representation, we propose physics-informed trajectory learning and integration into NeRF-based scene reconstruction.
We enable advanced obstacle handling through self-supervised scene decomposition and seamless integrated boundary constraints.
Our results showcase the ability to overcome challenges like occlusion uncertainty, density-color ambiguity, and static-dynamic entanglements.
Code and sample tests are at \url{https://github.com/19reborn/PICT_smoke}.
\end{abstract}

%
%


%
%


\begin{teaserfigure}
    \centering
    \vspace{-6pt}
    \includegraphics[width=\linewidth]{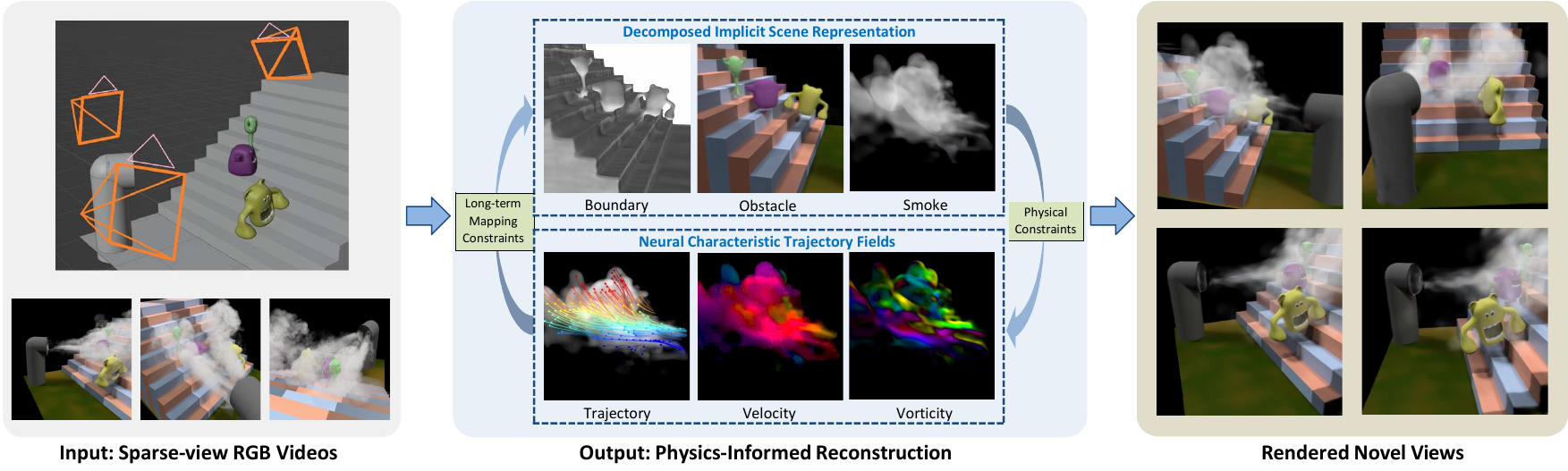}
    \vspace{-18pt}
    \caption{
    We present a physics-informed fluid reconstruction method using a novel Neural Characteristic Trajectory representation to preserve both short-term physics constraints and long-term conservation.
    In the challenging scene with smoke and obstacles, our method reconstructs decomposed radiance fields, obstacle geometry (serving as boundary constraints for smoke), smoke density, velocity, and trajectories from sparse-view RGB videos, and generates realistic renderings of novel views. }
    \label{fig:teaser}
\end{teaserfigure}
\vspace{2pt}

\maketitle
\input{tex/Sec1_Introduction}
\input{tex/Sec2_RelatedWork}
\input{tex/Sec3_Background}
\input{tex/Sec4_Lagrangian_field}

\input{tex/Sec6_Experiments}

\input{tex/Sec7_Conclusion}


\begin{acks}
{\small
We would like to thank the reviewers and the shepherd for their detailed and constructive comments. 
This work was supported in part by National Key R\&D Program of China 2022ZD0160802.
}
\vspace{-6pt}
\end{acks}
{\small
	\bibliographystyle{ACM-Reference-Format}
	\bibliography{egbib}
}

\clearpage
\appendix
\newpage
\setcounter{page}{1}

\input{tex/supp.tex}
\clearpage

\end{document}

%% file: tex/Sec1_Introduction.tex
\section{Introduction}

\begin{figure*}
	\includegraphics[width=0.96\linewidth]{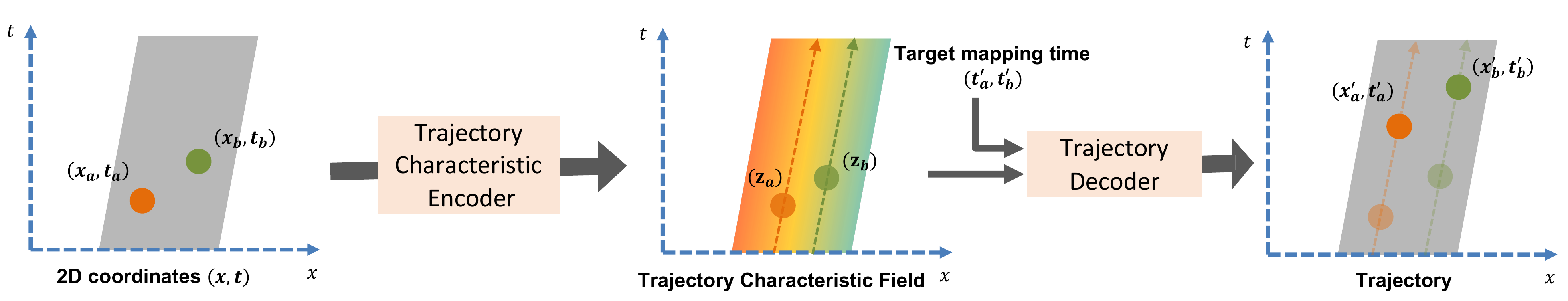}
 \vspace{-12pt}
	\caption
	{
        The Neural Characteristic Trajectory Field comprises a Trajectory Encoder extracting consistent features along trajectories and a Trajectory Decoder for trajectory prediction and velocity extraction via auto-differentiation.
	} 
	\label{fig:outline} 
\end{figure*}

Fluid dynamics fundamentally influences various phenomena, from aircraft navigation to sea circulation.
Understanding and predicting real-world fluid dynamics are crucial for climate forecasting~\cite{bi2023accurate}, vehicle design~\cite{rosset2023interactive}, medicine metabolism~\cite{sahli2020physics}, visual effects~\cite{chu2021learning}, and more. 
Despite its ubiquity and significance, accurately recovering flow from sparse observations remains challenging. 
We aim to reconstruct physically plausible fluids from sparse-view RGB videos using camera poses without relying on scene geometry or initial settings. It contributes to the goal of understanding fluids in the wild. 
Its complexity arises from limited observations of nonlinear dynamics, posing challenges, e.g., occlusion uncertainty, density-lighting ambiguity, and static-dynamic entanglement. 

For scene reconstruction, recent works~\cite{chu2022pinf, yu2023inferring} integrate Physics-Informed Neural Networks (PINN)~\cite{raissi2019pinn} with Neural Radiance Fields (NeRF)~\cite{mildenhall-ECCV2021-nerf} to facilitate partial differentiable equations (PDE) as end-to-end short-term constraints. 
However, the absence of long-term supervision renders them susceptible to local minima compromising image and local physics losses. 
Dynamic NeRFs~\cite{tretschk2020nonrigid, park2021hypernerf, park2020nerfies, pumarola2020dnerf} typically design deformation or scene flow fields for surface motion. These temporal connections drawn from visual cues are crucial in unifying multi-views but are inadequate for fluid and lack physics supervision.
In forward fluid simulation, advection with long-term conservation represents a significant research area. 
Many works~\cite{Qu:2019:bimocq, deng2023neural_flow_maps} 
follow the method of characteristic mapping (MCM)~\cite{wiggert1976original_mcm} to track flow maps for advection, achieving SOTA quality for vorticity preservation.
However, applying MCM to reconstruction is far from trivial, involving parameter-sensitive velocity integration in each training iteration and backward calculations for each integrated velocity.

To combine the potential of long-term and short-term physical constraints to mitigate uncertainty, 
we propose \textbf{Neural Characteristic Trajectory Fields}, a novel 
representation utilizing Eulerian neural fields to implicitly model the Lagrangian nature of fluid trajectories.
Illustrated in \myreffig{fig:outline}, we first use a trajectory encoder to transform spatio-temporal Eulerian coordinates into Lagrangian trajectory features. For a material particle at a given 4D coordinate, the output feature aims to capture sufficient information representing its entire spatiotemporal trajectory. 
Subsequently, a decoding network interprets the trajectory features at another time $t'$, generating the location $(x', y', z')$ the trajectory traverses at $t'$. 
In a scene-specific optimization, by constraining the trajectory feature to be consistent along its route, 
the encoder and decoder collectively serve as a closed-form velocity integral. Given a starting location $(x, y, z)$ and a time interval $(t,t')$, a single network forward pass generates the target location $(x', y', z')$ as the velocity integral.
Its advantages are threefold:
(i) Implicit neural trajectory enables velocity extraction via auto-differentiation, facilitating physics-informed training; 
(ii) Flow maps at arbitrary time intervals are obtained efficiently, avoiding time- and memory-intensive numerical integration; 
(iii) The forward trajectory calculation facilitates backward computations, enabling end-to-end optimization on long-term constraints without requiring differentiable solvers.

The primary distinction of our mapping, apart from widely used ones in deformable NeRFs~\cite{park2021hypernerf, park2020nerfies, pumarola2020dnerf}, or MCM in forward simulation~\cite{Qu:2019:bimocq, nabizadeh2022covector, deng2023neural_flow_maps}, lies in its spatiotemporally dense and continuous nature, making it topology-free and auto-differentiable. 
Specifically, it enables efficient forward inference of mappings and backward derivative calculations between any two points.
In contrast, existing mappings are often temporally sparse and pairwise, either confined to a canonical space~\cite{wang2023omnimotion} or necessitating frame-by-frame integration~\cite{Li2020NeuralSF}, both unsuitable in our case.

To derive the neural characteristic trajectories from sparse observations, we propose a physics-informed learning scheme that employs two categories of self-supervision.
The first set of \textbf{intrinsic constraints} directly arises from the Lagrangian transport essence of the trajectory. They regulate the representation and contribute to \textit{long-term} conservation of fluid dynamics. 
The second set of \textbf{physical constraints} is derived from PDEs and ensures the faithful behavior of fluid dynamics from an \textit{instantaneous perspective}.

Navigating to the application of smoke reconstruction, we seamlessly integrate our Neural Characteristic Trajectory with a spatial-temporal NeRF to obtain image supervision.  This integration features a unique \textit{dual-density} design, serving as a physically constrained neural radiance field for dynamic smoke.
Our trajectory further enhances reconstruction by imposing long-term constraints over the NeRF.  
For practical scenarios involving fluid-obstacle interactions, we propose a hybrid structure. This includes a Signed Distance Field (SDF) reconstruction for obstacles and a self-supervised decomposition that robustly separates fluid and obstacles without segmentation labels. 
The reconstructed SDF further provides critical boundary conditions for fluid dynamics, which were previously unattainable with rough volumetric obstacles.

To summarize, our work makes the following contributions:
\begin{itemize}\vspace{-4pt}
    \item 
    Introduction of the Neural Characteristic Trajectory Field, a novel hybrid Eulerian-Lagrangian representation, providing a closed-form automatic velocity integral . 
    This allows efficient flow map calculation between arbitrary frames, a capability not available previously.
    \item \vspace{-1pt}
    Integration of the trajectory field with a spatial-temporal NeRF for smoke reconstruction, showcasing improvements by concurrently leveraging long-term constraints and short-term physical priors.
    \item \vspace{-1pt}
    Presenting a self-supervised disentanglement of static obstacles and dynamic fluid in hybrid scenes. The high-fidelity obstacle geometry further provides crucial boundary constraints for fluid dynamics.
\end{itemize}\vspace{-6pt}
These contributions present advancements in facilitating high-fidelity fluid reconstruction in both synthetic and real scenes.

%% file: tex/Sec2_RelatedWork.tex
\section{Related Work}

\paragraph{Fluid Reconstruction} 
from observations has undergone extensive exploration. 
Established methods utilize active sensing with specialized hardware and lighting setups~\cite{10.1145/1409060.1409085,Gu13StructuredLight,Ji2013LightPath}.
Particle imaging velocimetry (PIV)~\cite{Grant97PIV,elsinga2006tomographic,xiong2017rainbow} tracks passive markers injected in the flow.
\citet{gregson2014capture} estimate velocity field from tomographic scanning and \citet{inproceedings} applies the Background Oriented Schlieren tomography (3D-BOS) to capture gas flows allowing the presence of occluding objects.
Recent advancements on RGB-video-based reconstruction reduce dependency on specialized setups.
To impose long-term temporal physics constraints for plausible reconstruction,
ScalarFlow~\cite{eckert2019scalarflow} proposes optimizing the ``residuals'' between reconstructed and simulated density and velocity jointly using a forward simulation solver. 
\citet{franz2021global} propose differentiable rendering and simulation to facilitate end-to-end optimization on density and velocity.
Both methods use limited volume resolutions with numerical errors for better efficiency.
They also assume known lighting conditions and lack support for scenes with obstacles. 
~\citet{deng2023vortex} infer vortex particles to represent and predict 2D fluid motion in videos; however, 3D cases are not supported.

\paragraph{The Method of Characteristic Mapping}
\cite{wiggert1976original_mcm} 
establishes long-term correspondence with flow maps.
Introduced to computer graphics by \citet{tessendorf2011characteristic}, MCM reduces numerical viscosity caused by velocity interpolations 
for forward simulation. 
While some approaches~\cite{sato2017long, sato2018spatially} trace virtual particles to calculate flow maps,
BiMocq$^2$~\cite{Qu:2019:bimocq} offers a better balance between efficiency and accuracy by preserving backward and forward maps with BFECC.
\citet{deng2023neural_flow_maps} use efficient neural representations of spatiotemporal velocity and excel in preserving velocity details and vorticity. 
Despite advancements in forward simulations, MCM lacks feasible differential calculations for neural reconstruction.

\paragraph{Implicit Neural Representation} 
(INR) is emerging as a continuous alternative to discretizations in modeling~\cite{park2019deepsdf, saito2019pifu, OccupancyNetworks, peng2020convolutional}, rendering~\cite{mildenhall-ECCV2021-nerf, barron-ICCV2021-mipnerf, muller-SIG2022-instantngp}, and simulation~\cite{chen2023implicit}.
The universal approximation theorem from the 1990s has proven its capability to represent highly nonlinear functions.
Advancements in various domains have empowered INRs to capture better high-frequency details with less memory footprint~\cite{tancik2020fourfeat, Chen2022ECCV, muller-SIG2022-instantngp, mihajlovic2023resfields}, provide well-defined higher-order differentials~\cite{sitzmann2020siren}, and perform efficient integral estimation~\cite{lindell2021autoint}.
Notable models like NeRF~\cite{mildenhall-ECCV2021-nerf} and PINN~\cite{raissi2019pinn} have adopted INR as closed-form estimation of radiance fields and physical fields with velocity and density, respectively. These models utilize the auto-diff of MLP for differentiable rendering and simulation, sparking a surge of research in rendering and physics. Our work contributes to INR-based physics-informed learning by supporting short-term physical priors and long-term temporal constraints simultaneously.

\paragraph{Learning Dynamic Scenes} 
with INR is typically accomplished by designing deformation fields~\cite{tretschk2020nonrigid,park2020nerfies,Li2020NeuralSF,park2021hypernerf}, scene flow fields~\cite{Li2020NeuralSF, du2021neuralRadianceFlow}, or trajectory~\cite{li-cvpr2023-Dynibar,wang2022neural, wang2021neuralTrajectory} 
to model the dense 3D motions of points according to visual cues.
In contrast, we focus on modeling the dynamic motion with a neural trajectory representation incorporated with physics, which has not received much attention yet.
PINF~\cite{chu2022pinf} and Hyfluid~\cite{yu2023inferring} are most relevant to ours. Combining PINN and NeRF for fluid reconstruction, these methods preserve short-term physics rules via soft constraints. However, these methods lack long-term supervision, and sparse observations are not fully utilized. Meanwhile, Hyfluid is not extended to scenes with obstacles. PINF reconstructs volumetric obstacles without addressing boundary conditions, whereas our work produces high-fidelity obstacle geometry, facilitating crucial boundary constraints for velocity reconstruction.

%% file: tex/Sec3_Background.tex
\section{Background Context} \label{sec:background}

We briefly summarize NeRF and PINN for fluid reconstruction.

\vspace{2pt}\noindent
\paragraph{Neural Radiance Field}
Given multi-view images with known poses, 
NeRF~\cite{mildenhall-ECCV2021-nerf} utilizes coordinate-based networks to represent a static scene as a continuous radiance field composed of density $\sigma (\mathbf{x}): \mathbb{R}^3 \to \mathbb{R}$ and color $\mathbf{c} (\mathbf{x}, \mathbf{v}):\mathbb{R}^3 \times \mathbb{S}^2 \to \mathbb{R}^3$, with position $\mathbf{x} \in \mathbb{R}^3$ and view direction $\mathbf{v} \in \mathbb{S}^2$ as inputs.
For each pixel, NeRF sample $n$ points along its camera ray $\mathbf{r}$ with center $\mathbf{o}$ and direction $\mathbf{v}$ using $\{{\mathbf{r}(h_i)}=\mathbf{o}+h_i\mathbf{v}|i=0,1,\dots, n-1\}$.
Through discretized radiance accumulation, pixel colors $\hat{\mathbf{C}}$ are computed as
\resizeEq{
\hat{\mathbf{C}}(\mathbf{r}) =\sum_{i=1}^N T_i\Bigl(1-\exp (-\sigma_i \delta_i)\Bigr) \mathbf{c}_i 
\text{,~} 
T_i =\exp \Bigl(-\sum_{j=1}^{i-1} \sigma_j \delta_j\Bigr)
\text{,~} 
\delta_j = h_{j+1} - h_{j} .
}{eqn:vol_rendering_discreted}{0.92\linewidth}
With the differentiable calculations, NeRF simultaneously optimizes two MLPs, a “coarse” one as a probability density
function for importance sampling and a “fine” one as the target radiance field  $(\sigma, \mathbf{c})$ using the loss {\small $\mathcal{L}_{image}=\sum_{\mathbf{r} \in \mathcal{R}}\left\|\hat{\mathbf{C}}_c(\mathbf{r})-\mathbf{C}(\mathbf{r})\right\|_2^2 + \left\|\hat{\mathbf{C}}_f(\mathbf{r})-\mathbf{C}(\mathbf{r})\right\|_2^2$}, where $\hat{\mathbf{C}}_c, \hat{\mathbf{C}}_f$ and $\mathcal{R}$ stand for predicted colors of the coarse and fine models and the set of rays in each batch, respectively.

\input{fig/illustration/method_overview}
\vspace{4pt}\noindent
\paragraph{Physics-Informed Velocity Estimation} 
Fluids follow the well-known incompressible Navier-Stokes equations:
\begin{equation}
\label{eqn:NS_equation}
\frac{\partial \mathbf{u}}{\partial t}+\mathbf{u} \cdot \nabla \mathbf{u} =-\frac{1}{\rho} \nabla p+\nu \nabla \cdot \nabla \mathbf{u}+\mathbf{f} \text{~~~~and~~~} 
\nabla \cdot \mathbf{u} =0 ,
\end{equation}
where $\boldsymbol{u},p,\rho,\nu,$ and $\boldsymbol{f}$ represent velocity, pressure, density, the dynamic viscosity coefficient, and the external force respectively.
Meanwhile, density evolution follows the transport equation: $\frac{\partial d}{\partial t} + \mathbf{u} \cdot \nabla d = 0 $.
According to the Beer-Lambert law, concentration density $d$ can be considered proportional to the optical density $\sigma$.
By extending a static NeRF to a time-varying {\small$\mathcal{F}_{NeRF}(\mathbf{x},t)=(\sigma, \mathbf{c})$} and combining it with a spatiotemporal velocity network {\small $\mathcal{F}_{PINN}(\mathbf{x},t)=\mathbf{u}$}, PINF\cite{chu2022pinf} applies the image loss $\mathcal{L}_{image}$ to estimate density from videos and simplified physical priors as follows to seek feasible velocity solutions using the density field with minimal impact from extra forces, pressure differences, and viscosity:
\resizeEq{
\mathcal{L}_{\frac{D \sigma}{D t}} =\left(\frac{\partial \sigma}{\partial t}+\mathbf{u} \cdot \nabla \sigma\right)^2
 \text{~~and~} 
\mathcal{L}_{N S E} =\left\|\frac{\partial \mathbf{u}}{\partial t}+\mathbf{u} \cdot \nabla \mathbf{u}\right\|_2^2+w_{d i v}\|\nabla \cdot \mathbf{u}\|_2^2 .
}{eqn:ns_loss}{0.92\linewidth}
While producing reasonable results, PINF overlooks long-term conservations, resulting in smooth and dissipative solutions.

%% file: fig/illustration/method_overview.tex
%
%
\begin{figure*}
\begin{minipage}{0.25\textwidth}
{ \footnotesize \setlength\tabcolsep{3pt}
\begin{tabularx}{\linewidth}{clX}
\toprule[1pt]
Symbol & Type & Def. \\\hline
$ \mathbf{x}$ & vector & Spatial position\\\hline
$t$           & scalar & Time\\\hline
$\mathbf{z}$ & vector &  Trajectory feature\\\hline
$\mathcal{E}$ & network & Trajectory encoder \\\hline
$\mathcal{D}$  & network  & Trajectory decoder \\\hline
$\mathbf{x}(t)$ & function & Particle trajectory\\\hline
$\mathbf{u}(\mathbf{x},t)$ & function & Eulerian velocity field, $\mathbb{R}^3 \times \mathbb{R} \to \mathbb{R}^3$\\\hline
$\hat{\mathbf{u}}(\mathbf{z},t)$ & function & Lagrangian velocity field, $\mathbb{R}^D \times \mathbb{R} \to \mathbb{R}^3$\\\hline
\end{tabularx}
 }
 \\\vfill
 \centering
 \includegraphics[width=0.7\linewidth]{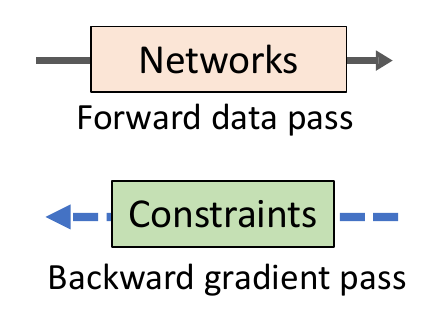}
\end{minipage}
\hfill
 \begin{minipage}{0.72\textwidth}
\includegraphics[width=\linewidth]{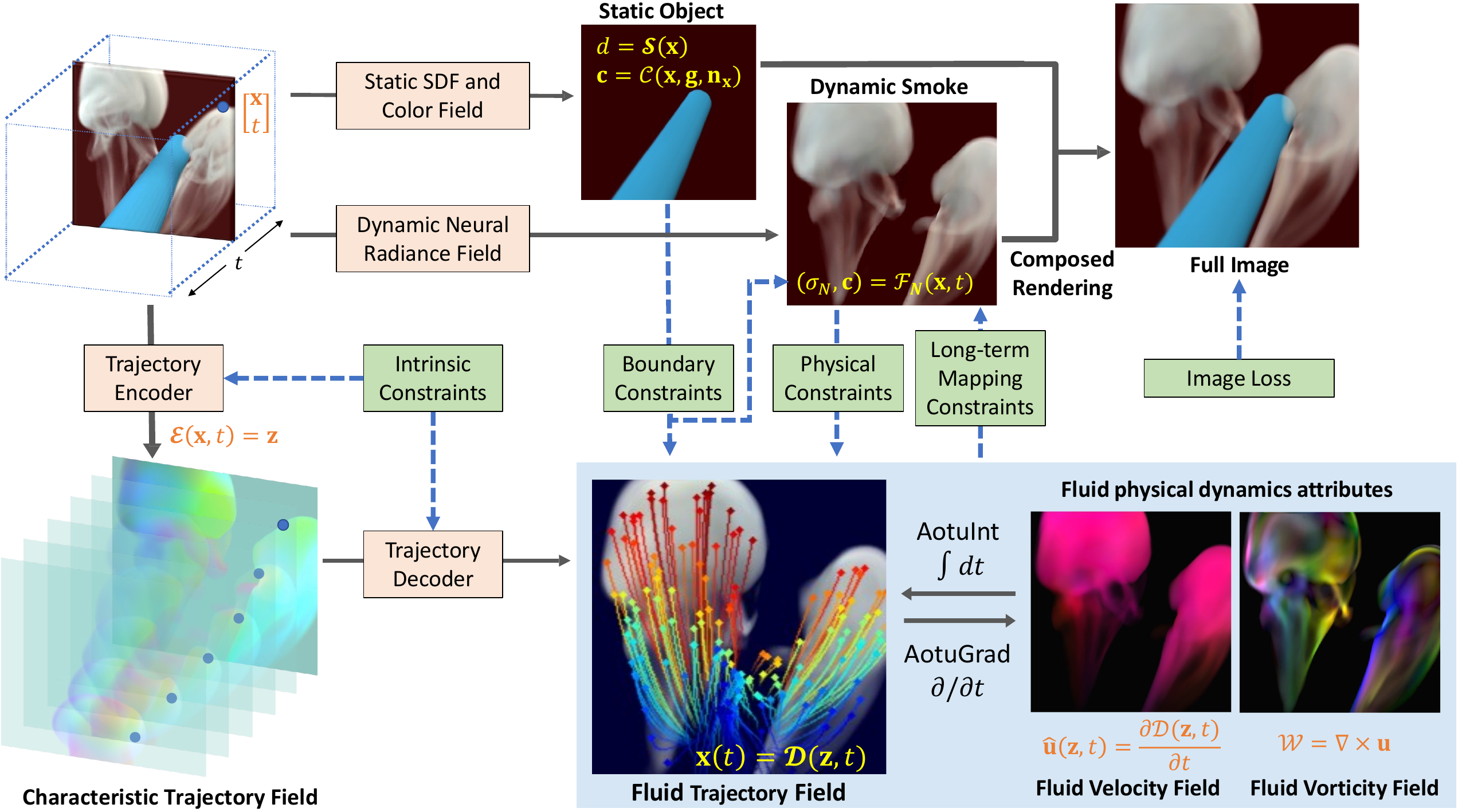}
\end{minipage}
	\caption
	{
        Method overview. We model static objects using SDF and color fields, and dynamic smoke using a time-varying NeRF. 
        The Neural Characteristic Trajectory Field enables end-to-end optimization of long-term mapping constraints and short-term physics priors. 
        We utilized boundary conditions extracted from SDF-based obstacle geometry.
	}
	\label{fig:method_overview}
	%
\end{figure*}
%
%

%% file: tex/Sec4_Lagrangian_field.tex
\section{Reconstruction with Neural Characteristic Trajectory Field}

In this section, we present the Neural Characteristic Trajectory Field (Sec~\ref{sec:NCTJ}), discuss its physics-informed learning (Sec~\ref{sec:learn_NCTJ}), integrate it into NeRF-based reconstruction (Sec~\ref{sec:intergration_with_nerf}), and handle obstacles (Sec~\ref{sec:boundary_recon}). Fig. \ref{fig:method_overview} offers an overview of our full method, illustrating the connections between all components and summarizing the comprehensive supervision facilitated by our trajectory field.

\subsection{Neural Characteristic Trajectory Field} \label{sec:NCTJ}
Fluid dynamics is approached as particle motions from a Lagrangian perspective.
Assume each particle's trajectory has its own characteristic, which 
can be abstracted as a high-level embedding $\mathbf{z} \in \mathbb{R}^D$, then all trajectory embeddings form a continuous spatial-temporal function. We use a Multi-Layered Perceptron (MLP) $\mathcal{E}$ to learn an encoding of this function: $\mathbf{z} = \mathcal{E}(\mathbf{x}, t)$ with $\mathbf{x}\in \mathbb{R}^3 $ and $t \in \mathbb{R}$.
Then, network $\mathcal{E}$ establishes connections between Eulerian and Lagrangian representations.
In practice, we use $D=16$, and more details on network structures can be found in the supplementary.

We visualize an example trajectory feature field in Fig.~\ref{fig:illustration_trajectory_feature}, confirming our assumption that particles' consistent features along trajectories form a continuous field.
\input{fig/illustration/illustration_trajectory_feature}
Any feature $\mathbf{z}$ sampled from this neural Field corresponds to a specific particle path $\mathbf{x}(t)$.
We use another MLP-based network $\mathcal{D}$ to decode this path: $\mathbf{x}(t) = \mathcal{D}(\mathbf{z}, t)$.

In a scene-specific optimization, the Neural Characteristic Trajectory Field constituted by $\mathcal{E}$ and $\mathcal{D}$ can rapidly predict the temporal evolution of arbitrary points.
Given a 4D point $(\mathbf{x},t)$ and another time $t'$, we can directly get its corresponding position $\mathbf{x}'$ at time $t'$ through a single feed-forward pass: $\mathbf{x}' = \mathcal{D}(\mathcal{E}(\mathbf{x},t), t')$.
Compared to methods \cite{chu2022pinf} that establish temporal correspondences by warping with velocity integrated over up to two frames, our representation requires only a single feed-forward computation while supporting long-term mappings even beyond 50 frames. 
This enables us to impose long-term mapping constraints essential for high-fidelity reconstruction.
For a spatial-temporal velocity field $\mathbf{u}(\mathbf{x},\tau):\mathbb{R}^3 \times \mathbb{R} \to \mathbb{R}^3$ 
, our approach offers a differentiable closed-form calculation of its integral over the time $\tau$:
\begin{equation}\small
    \begin{aligned} \label{eqn:integral_velocity}
        \int_{t}^{t'} \mathbf{u} d\tau = \mathbf{x}' - \mathbf{x} = \mathcal{D}(\mathcal{E}(\mathbf{x},t), t') - \mathcal{D}(\mathcal{E}(\mathbf{x},t), t), 
    \end{aligned}
\end{equation}
eliminating infeasible requirements for frame-by-frame velocity backpropagation during each training iteration.

\subsection{Physics-Informed Learning} \label{sec:learn_NCTJ}
\input{tex/Sec4_2_Physics_Informed}

\subsection{Integration with Neural Radiance Field} \label{sec:intergration_with_nerf}
\input{tex/Sec4_3_NeRF_integration}

\subsection{Reconstruction of Scenes with Obstacles} \label{sec:boundary_recon}

\input{tex/Sec4_4_obstacle}

%% file: fig/illustration/illustration_trajectory_feature.tex
%
%
\begin{figure}[b]
	\includegraphics[width=0.9\linewidth]{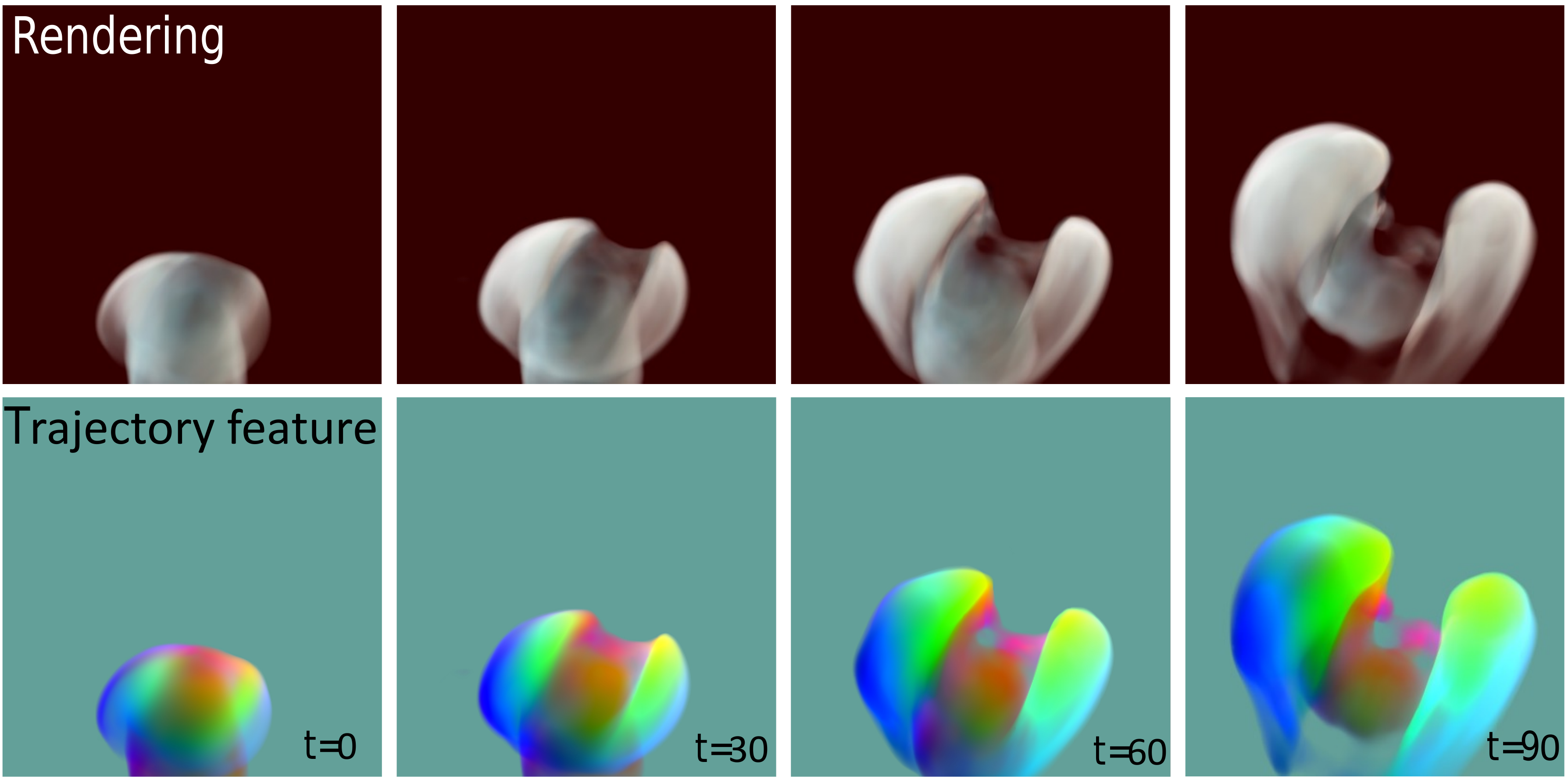}  
	\caption
	{
		Illustration of trajectory feature in the neural trajectory field. 
        Features along specific trajectories remain consistent. 
        We use the first three dimensions of the trajectory feature and normalize them based on the maximum and minimum values across sequences to visualize the color.
	}
	\label{fig:illustration_trajectory_feature}
	%
\end{figure}
%
%

%% file: tex/Sec4_2_Physics_Informed.tex
While our proposed representation holds great potential, reconstructing it solely from visual cues is highly challenging for fluids.
Merely including PDE-based physics priors is also not enough, as they are local constraints and permit error accumulations over the long term. 
Hence,  we first introduce a set of self-supervised constraints termed \textit{Intrinsic Constraints}.
They align with the Lagrangian essence of the trajectory and play crucial roles in effectively regularizing $\mathcal{E}$ and $\mathcal{D}$ during training.

\noindent\paragraph{Intrinsic Constraints}
The first constraint $\mathcal{L}_{cycle}^{self}$ is straight-forward, stating that a particle at location $\mathbf{x}$ and time $t$ should be mapped back to itself with the same time $t$: 
\begin{equation}
    \begin{aligned} \label{Loss:self_cycle_loss}
        \mathcal{L}_{cycle}^{self} = \lVert \mathcal{D}(\mathcal{E}(\mathbf{x},t), t) - \mathbf{x} \rVert_2^2 .
    \end{aligned}
\end{equation}
In addition to local mapping, we incorporate long-term consistency using 
the Neural Characteristic Trajectory Field:
\begin{subequations}
\begin{align}
\mathcal{L}_{cycle}^{cross}&= \lVert \mathcal{E}(\mathbf{x}_a, t_a) - \mathcal{E}(\mathbf{x}_b, t_b) \rVert_2^2 \label{Loss:cross_cycle_loss} \\\vspace{-2pt}
\text{~~with~~} \mathbf{x}_b = & \mathbf{x}_a + \mathcal{D}(\mathcal{E}(\mathbf{x}_a,t_a), t_b) - \mathcal{D}(\mathcal{E}(\mathbf{x}_a,t_a), t_a) . \label{eqn:long_mapping} 
\end{align}
\end{subequations}
We uniformly sample $(\mathbf{x_a}, t_a)$ and $t_b\sim$ Uniform$(t_a-\Delta t, t_a+\Delta t)$. Interval $\Delta t$ progressively increases during training.
Derived from Eq.~\ref{eqn:integral_velocity}, Eq.~\ref{eqn:long_mapping} is slightly more accurate than directly using $\mathcal{D}(\mathcal{E}(\mathbf{x}_a,t_a), t_b)$ when $\mathcal{L}_{cycle}^{self}$ is not neglectable at the start of training.

As a passive quantity flowing without diffusion, the trajectory feature adhere to the transport equation:
$\frac{\partial \mathbf{z}}{\partial t} + \mathbf{u} \cdot \nabla \mathbf{z} = \mathbf{z}_\text{source}.$
Since the source term $\mathbf{z}_\text{source}$ is nonzero only in few locations with inflow or outflow fluid, we formulate a loss $\mathcal{L}_{feature}$:
\begin{equation}
    \begin{aligned} \label{Loss::trajectory_feature_material_derivative} \scriptsize
        \mathcal{L}_{feature} & = \left\lVert \frac{\partial \mathcal{E}(\mathbf{x},t)}{\partial t}+\mathbf{u} \cdot \nabla \mathcal{E} (\mathbf{x},t) \right\rVert_2^2 .
    \end{aligned}
\end{equation}
The calculation of $\mathbf{u}$ is given in the next paragragh.
All intrinsic constraints, $\mathcal{L}_{cycle}^{self}$, $\mathcal{L}_{cycle}^{cross}$ and $\mathcal{L}_{feature}$, collectively contribute to the core assumption making our Neural Characteristic Trajectory Field
theoretically sound, 
which is that the trajectory feature remains the same along its pathline. 
While intrinsic constraints regularize this Lagrangian essence, \textit{Physical Constraints} are required so that motions reflect fluid dynamics.

\noindent\paragraph{Physical Constraints} 
Given an arbitrary 4D point $(\mathbf{x}, t)$ and its corresponding trajectory feature $\mathbf{z}$ obtained from $\mathcal{E}$, its Eulerian velocity $\mathbf{u}(\mathbf{x},t)$ equals to its Lagrangian velocity $\hat{\mathbf{u}}(\mathbf{z},t)$, i.e., $\mathbf{u}(\mathbf{x},t)=\hat{\mathbf{u}}(\mathcal{E}(\mathbf{x}, t),t)=\hat{\mathbf{u}}(\mathbf{z},t)$.
These velocity values can be derived by taking the partial derivative of the Neural Characteristic Trajectory Field with respect to the time $t$:
\begin{equation} 
\label{eqn:definition_velocity}
    \begin{aligned} 
         \mathbf{u}(\mathbf{x},t)=\hat{\mathbf{u}}(\mathbf{z}, t) =\frac{\partial \mathcal{D}(\mathbf{z}, t)}{\partial t}=\frac{\partial \mathcal{D}(\mathcal{E}(\mathbf{x}, t_1), t_2)}{\partial t_2},
    \end{aligned}
\end{equation}
where $t_1, t_2$ are the time inputs of $\mathcal{E}$ and $\mathcal{D}$ that share the same time
value consistently. 
The rationale we only take partial derivatives with respect to $t_2$ but not $t_1$ is because, when a particle moves along its own trajectory, its feature $\mathbf{z}=\mathcal{E}(\mathbf{x}, t_1)$ should be unchanged.
Thus, by maintaining separate backward gradients for $t_1$ and $t_2$, we can compute the function of $\hat{\mathbf{u}}$ via auto-differentiation of the network $\mathcal{D}$. The material velocity derivative $\frac{D\mathbf{u}}{Dt}$ can be calculated as $\frac{\partial \hat{\mathbf{u}}(\mathbf{z}, t)}{\partial t}$ via the 2nd-order auto-differentiation of $\mathcal{D}$ accordingly.

We integrate physical constraints from the transport equation and the Navier-Stokes equations, under same assumptions as PINF \cite{chu2022pinf} which seeks feasible solutions with minimal impact from extra forces, pressure differences, and viscosity:
\resizeEq{
\mathcal{L}_{transport} =\left(\frac{\partial \sigma}{\partial t}+\hat{\mathbf{u}} \cdot \nabla \sigma\right)^2  \text{~~,~~~} \mathcal{L}_{NSE} & = \left |\frac{D\mathbf{u}}{Dt}\right | = \left |\frac{\partial \hat{\mathbf{u}}}{\partial t}\right|= \left|\frac{\partial^2 \mathcal{D}(\mathbf{z}, t)}{\partial t^2}\right| ,\\
\text{~~and~~} \mathcal{L}_{div}  = \Big|\nabla \cdot \hat{\mathbf{u}}\Big| & =\Big|\nabla \cdot \frac{\partial \mathcal{D}(\mathcal{E}(\mathbf{x}, t_1), t_2)}{\partial t_2}\Big|,
}{Loss:physical_constraints}{0.85\linewidth}
with the fluid density $\sigma$ to be introduced in Sec~\ref{sec:intergration_with_nerf}.

Besides short-term physical constraints, the trajectory field also facilitates long-term velocity constraints.
Akin to forward simulations, covector advection~\cite{nabizadeh2022covector} contributes to globally plausible and non-dissipating velocity evolution. For uniformly sampled $(\mathbf{x_a}, t_a)$ and $t_b \sim${Uniform}$(t_a-\Delta t, t_a+\Delta t)$, we propose:
\begin{equation}
    \begin{aligned}
    \mathcal{L}_{mapping}^{velocity} = |\hat{\mathbf{u}}_a - (M_{a\rightarrow b})^T \hat{\mathbf{u}}_b| , ~~\text{with}~~\\
\mathbf{z}_a = \mathcal{E}(\mathbf{x_a}, t_a), ~~ \hat{\mathbf{u}}_a = \hat{\mathbf{u}} (\mathbf{z}, t_a), ~~ \hat{\mathbf{u}}_b = \hat{\mathbf{u}} (\mathbf{z}, t_b) , \\
~~\text{and}~~ M_{a\rightarrow b} = \frac{\partial \mathcal{D}( \mathcal{E}(\mathbf{x}_a, t_a), t_b)}{\partial \mathbf{x}_a}.\\
    \end{aligned} \label{Loss:velocity_mapping_loss}
\end{equation}
following the idea of covector advection as ${\mathbf{u}}_a=(M_{a\rightarrow b})^T {\mathbf{u}}_b$.

\noindent\paragraph{Equivalence to Eulerian Advection} 
While $\mathcal{L}_{NSE}$ in Eq.~\ref{Loss:physical_constraints} addresses advection loss from a Lagrangian perspective, 
it is equivalent to the Eulerian alternative employed in previous work. 
Given 
\begin{equation}\label{Loss:velocity_eulerian_derivative}
     \frac{\partial \mathbf{u}(\mathbf{x},t)}{\partial t } =\frac{\partial \hat{\mathbf{u}}(\mathbf{z},t) }{\partial \mathbf{z}} \cdot \frac{\partial \mathbf{z}(\mathbf{x},t) }{\partial t} + \frac{\partial \hat{\mathbf{u}} }{\partial t}  \text{~~ and ~~}
 (\mathbf{u} \cdot \nabla) \hat{\mathbf{u}}  = \frac{\partial \hat{\mathbf{u}} }{\partial \mathbf{z}}   (\mathbf{u} \cdot \nabla) \mathbf{z},
\end{equation}
the Eulerian advection term expresses as
 \begin{equation} \label{Loss:velocity_eulerian_advection}
    \frac{\partial \mathbf{u} }{\partial t } +  (\mathbf{u} \cdot \nabla) \mathbf{u}  = \frac{\partial \hat{\mathbf{u}} }{\partial \mathbf{z}} \cdot \left(\frac{\partial \mathbf{z} }{\partial t} + (\mathbf{u} \cdot \nabla) \mathbf{z} \right)  + \frac{\partial \hat{\mathbf{u}} }{\partial t} ,
 \end{equation}
which is effectively penalized through applying $\mathcal{L}_{feature}$ and $\mathcal{L}_{NSE}$ in Eq.~\ref{Loss:physical_constraints}.
Utilizing $\mathcal{L}_{feature}$ and $\mathcal{L}_{NSE}$ instead of an Eulerian advection loss offers the advantage of limiting the scale for auto-differentiation, offering improved efficiency.

%% file: tex/Sec4_3_NeRF_integration.tex
While a time-varying NeRF $\mathcal{F}_{N} (\mathbf{x},t) = (\sigma_{N}, \mathbf{c})$ can represent a dynamic scene, 
training it with sparse views becomes highly underdetermined, as extensively discussed~\cite{Gao-ICCV21-DynNeRF, gafni2020dynamic}.
This underdetermined density
further contaminates the velocity through $\mathcal{L}_{transport}$.
To mitigate the underdetermined nature, we propose a novel dual-density strategy complemented by long-term temporal supervision using our trajectory field.

Previous methods often employ a coarse-and-fine strategy, simultaneously optimizing a coarse Siamese NeRF for importance sampling when learning the fine NeRF~\cite{mildenhall-ECCV2021-nerf, chu2022pinf}.
Our dual-density strategy retains the fine model $\mathcal{F}_{N} (\mathbf{x},t)$ as an Eulerian representation, 
and replaces the coarse one with an MLP-based Lagrangian density model $\sigma_{L} = \mathcal{F}_{L} (\mathcal{E}(\mathbf{x},t))$.
This Lagrangian density $\sigma_{L}$ takes our trajectory features as a prior input and is inherently more consistent along particle trajectories.

We train $\mathcal{F}_{N}$ with a combination of image and mapping losses:
\resizeEq{
\mathcal{L}_{img}=\sum_{\mathbf{r} \in \mathcal{R}}\left\|\hat{C}(\mathbf{r})-C(\mathbf{r})\right\|_2^2,
\mathcal{L}_{mapping}^{density} = |\sigma_{a} - \sigma_{b}| 
& \text{~~and~~}
\mathcal{L}_{mapping}^{color} = |\mathbf{c}_{a} - \mathbf{c}_{b}|\\
\text{~~with~~}  (\sigma_{a}, \mathbf{c}_{a})  = \mathcal{F}_{N}(\mathbf{x}_a,t_a) ,
& (\sigma_{b}, \mathbf{c}_{b})   = \mathcal{F}_{N}(\mathbf{x}_b, t_b)
}{Loss:density_mapping_loss}{0.9\linewidth}

The sampling of $(\mathbf{x}_a,t_a)$, $t_b$ and the calculation of $\mathbf{x}_b$ are consistent with those in $\mathcal{L}_{cycle}^{cross}$.
Akin to knowledge distillation, we use the NeRF density $\sigma_{N}$ as the teacher to train the Lagrangian density $\sigma_{L}$ with $\mathcal{L}_{distillation} = |\sigma_{N} - \sigma_{L}|$. 
Note that this loss's gradient influences the trajectory encoder $\mathcal{E}$, improving feature identification implicitly based on density continuity.

Compared with $\mathcal{F}_{N}$, the Lagrangian density has limited degrees of freedom and is strongly constrained by physics.
To illustrate the difference, we visualize $\sigma_{N}$ and  $\sigma_{L}$ in Fig.~\ref{fig:illustration_two_layer_density}.
\input{fig/illustration/illustration_two_layer_density}
The Lagrangian density $\sigma_{L}$ exhibits gradients that are significantly more structured, making it well-suited for training the velocity using transport loss. 
The NeRF density $\sigma_{N}$ effectively captures more high-frequency details, which is instrumental in reconstructing high-fidelity density and velocity results.
Thus, our dual-density strategy applies the original NeRF density $\sigma_{N}$ with larger DOFs as scene representations, while employing the better-constrained $\sigma_{L}$ as velocity guidance 
via
\begin{equation}
    \begin{aligned} \label{Loss:full_density_transport}
        \mathcal{L}_{transport}^{full}  &= \left(  \frac{\partial \sigma_{L}}{\partial t}  + \mathbf{u} \cdot \nabla \sigma_{L}  \right)^2 + \lambda \left(  \frac{\partial \sigma_{N}}{\partial t} + \mathbf{u} \cdot \nabla \sigma_{N}  \right)^2 .
    \end{aligned}
\end{equation}
This allows the velocity field to learn robustly a structural ansatz from the Lagrangian density $\sigma_{L}$ and to gather detailed information from the NeRF density $\sigma_{N}$. We also conduct an ablation study evaluating the dual-density representation in the supplementary.

%% file: fig/illustration/illustration_two_layer_density.tex
%
%
\begin{figure}
	%
	\includegraphics[width=.95\linewidth]{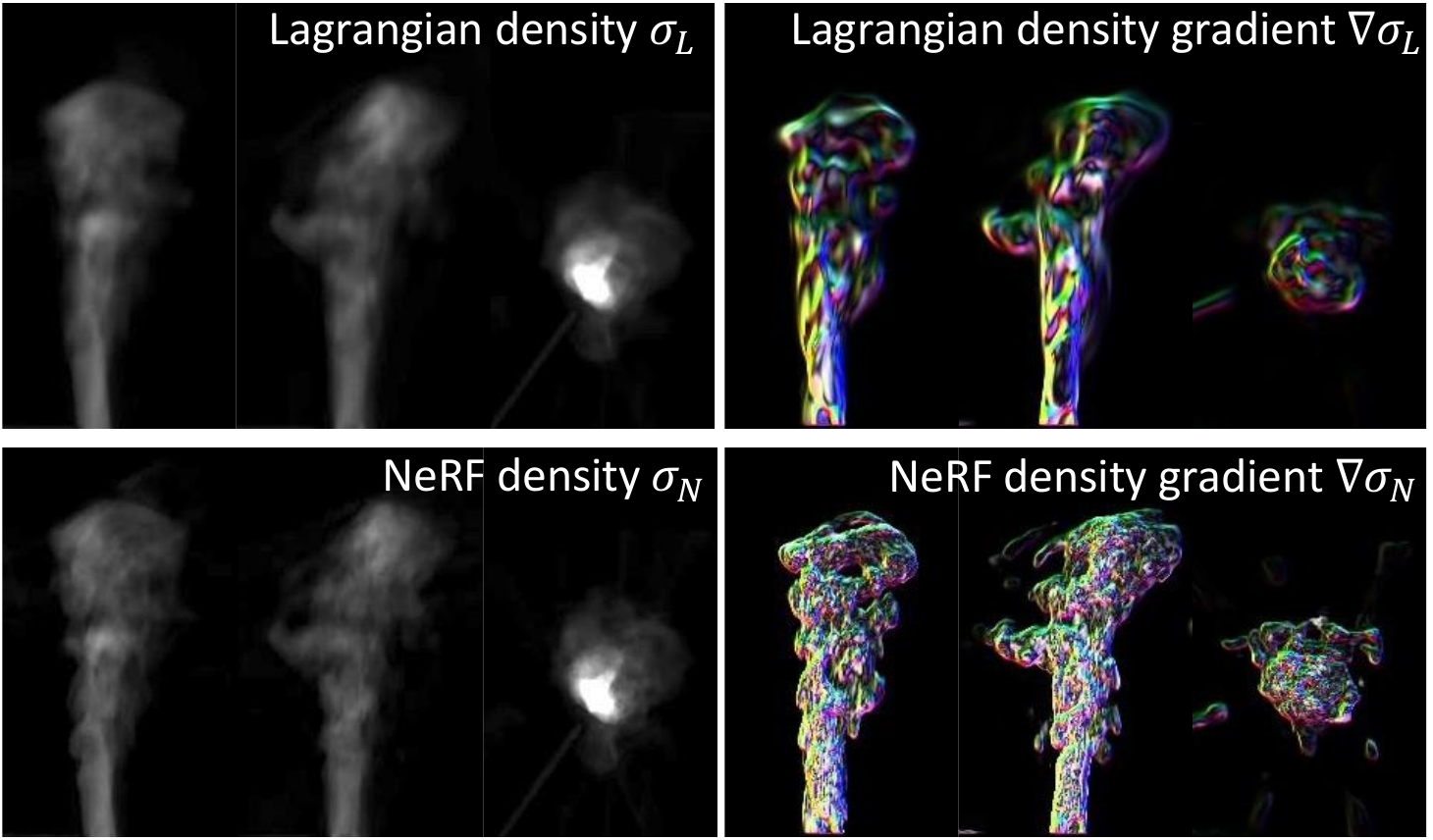}  
        \vspace{-12pt}
	\caption
	{
		Front-side-top illustrations of Lagrangian and NeRF density. 
        The Lagrangian density yields more robust and smoother outcomes in the reconstructed density and its gradients, whereas the NeRF density contains richer high-frequency details.
	}
	\label{fig:illustration_two_layer_density}
	%
\end{figure}
%
%

%% file: tex/Sec4_4_obstacle.tex
\paragraph{SDF-based boundary conditions} 
PINF~\cite{chu2022pinf} enables the joint reconstruction of fluid and obstacle for the first time, utilizing a time-varying NeRF for dynamic components and a vanilla NeRF for static components in a hybrid scene. However, boundary conditions were still ignored since the extraction of high-quality surfaces is hindered by the volumetric representation.
Together with our dual-density fluid representation, we utilize NeuS~\cite{wang-nips2021-neus} to represent static obstacles as a signed distance field (SDF) facilitating the introduction of critical boundary constraints.

We use two MLPs, ${\mathcal{S}}$ and ${\mathcal{C}}$, to model SDF value $s$ and color $\mathbf{c}$ as 
\begin{equation} \label{eqn:neus_sdf_color}
    {\mathcal{S}}(\mathbf{x}) = (s, \mathbf{g}) \text{~~~~ and ~~~~}
    {\mathcal{C}}(\mathbf{x}, \mathbf{g}, \vect{n}_{\mathbf{x}}) = \mathbf{c}.
\end{equation}
$\mathcal{C}$ considers the position $\mathbf{x}$, the geometry feature $\mathbf{g}$ obtained from $\mathcal{S}$, as well as the normal vector of the implicit surface $\vect{n}_{{x}}=\nabla_{\mathbf{x}}{\mathcal{S}}(\mathbf{x})$.
Following \citet{yariv2020Iigr}, 
we apply the loss $\mathcal{L}_{eikonal}=\frac{1}{m} $ $\sum_{i}\left(\left|\nabla_{\mathbf{x}} \mathbf{s}_i \right|-1\right)^2$ for the Eikonal equation $\left|\nabla_{\mathbf{x}} \mathbf{s}\right| = 1$, with $m$ being the total number of the sampled points during volume rendering.

Our boundary constraints dictate the absence of smoke within the interior of static objects, 
and fluids should neither enter nor exit the solid surface, i.e., the fluid velocity should satisfy: $\mathbf{u} \cdot \vect{n}_{\mathbf{x}} = 0$.
Leveraging our SDF representation, we can directly assess whether points lie within the interior of an object, a capability previously unattainable. 
We then convert these priors as loss constraints for our Neural Characteristic Trajectory Field as

\input{fig/illustration/illustration_reconstruction}
\input{fig/comparison/fig_decomposed_novel_view_comp}

\begin{equation} 
    \begin{split} \label{eqn:vel_boundary_loss}
    \mathcal{L}_{bnd}^{\sigma} &=  \sum_{\mathbf{x}} (\mathcal{S}(x) \leq 0) \cdot  \sigma(x) \\
        \mathcal{L}_{bnd}^{\mathcal{U}} &=  \sum_{x} \Big( \underbrace{ (|\mathcal{S}(x)| \le \epsilon ) \left\| \mathbf{u}(x,t)  \cdot \hat{n}\right\|}_{\text{surface constraint}}
        + \underbrace{ (\mathcal{S}(x) < 0 ) \left\| \mathbf{u}(x,t) \right\|}_{\text{internal constraint}} \Big) .
    \end{split}
\end{equation}
A negative SDF, $\mathcal{S}(\mathbf{x}) < 0$, indicates that position $\mathbf{x}$ is inside the obstacle. In practice, we choose $\epsilon = 0.01$.
%

\paragraph{Self-supervised scene decomposition}
The key cause of the static-dynamic entanglement is that the static part can be easily modeled by the dynamic model.
To address this, we first adopt the layer-by-layer growing strategy~\cite{chu2022pinf} to prevent the overfitting of our dynamic model at the beginning of training. 
Then, we propose improvements on image supervision.
Previous work~\cite{chu2022pinf} merely constrains the composite rendering of static and dynamic parts with
$\mathcal{L}_{img}^{composite} = \sum_{\mathbf{r} \in \mathcal{R}} \left\|\hat{C}_{composite}(\mathbf{r})-
    C(\mathbf{r})\right\|_2^2$,
with $\hat{C}_{composite}$ calculated from density samples combining smoke and obstacle.
More details regarding our composite rendering of an SDF surface and a NeRF field can be found in the supplement (Supp.~\ref{sec:hybrid_render}).
We propose to further treat dynamic elements as a ``translucent occlusion'' and explicitly require the static model to reconstruct persistent parts from ``occluded'' images.
This shift aligns our task with the in-the-wild NeRF reconstruction~\cite{martin2021nerf_in_the_wild}.
Accordingly, we introduce the occluded image loss:
\begin{equation}
    \begin{aligned} 
    \mathcal{L}_{img}^{occluded} &= \sum_{\mathbf{r} \in \mathcal{R}} (\mathcal{V}_r \left\|\hat{C}_s(\mathbf{r})-
    C(\mathbf{r})\right\|_2^2 + \lambda \left(1-\mathcal{V}_r\right)^2 ) ,\\
        \mathcal{V}_r &= 1 - \sum_{i=1}^N T_i\left(1-\exp \left(-\sigma_i \delta_i\right)\right) .
    \end{aligned}
\end{equation}
Here, $\hat{C}_s(\mathbf{r})$ is the rendered color of the static model and $\mathcal{V}_r$ is the visibility of the static scene calculated using our fluid density $\sigma_N$.
We sample $\sigma_i$ from $\sigma_N$ and $T_i$ accumulates $\sigma_i$.
Higher visibility values emphasize pixel importance, assuming association with static components in our case. 
This term is balanced by the second term in $\mathcal{L}_{img}^{occluded}$, 
serving as a regularization to encourage the capture of persistent radiance in the static model as much as possible. We employ $\mathcal{L}_{img}^{occluded}$ at the beginning of the training process and progressively transit to $\mathcal{L}_{img}^{combined}$ using
\begin{equation}
    \begin{aligned} \label{Loss:final_combined_rgb_loss}
    \mathcal{L}_{img}^{combined} = (1-\alpha) \mathcal{L}_{img}^{occluded} + \alpha \mathcal{L}_{img}^{composite} .
    \end{aligned}
\end{equation}
This self-supervision ensures the effective reconstruction of the static scene amidst dynamic entanglement, as illustrated in Fig.~\ref{fig:illustration_reconstruction}.

%% file: fig/illustration/illustration_reconstruction.tex
%
%
\begin{figure}
	\includegraphics[width=0.96\linewidth]{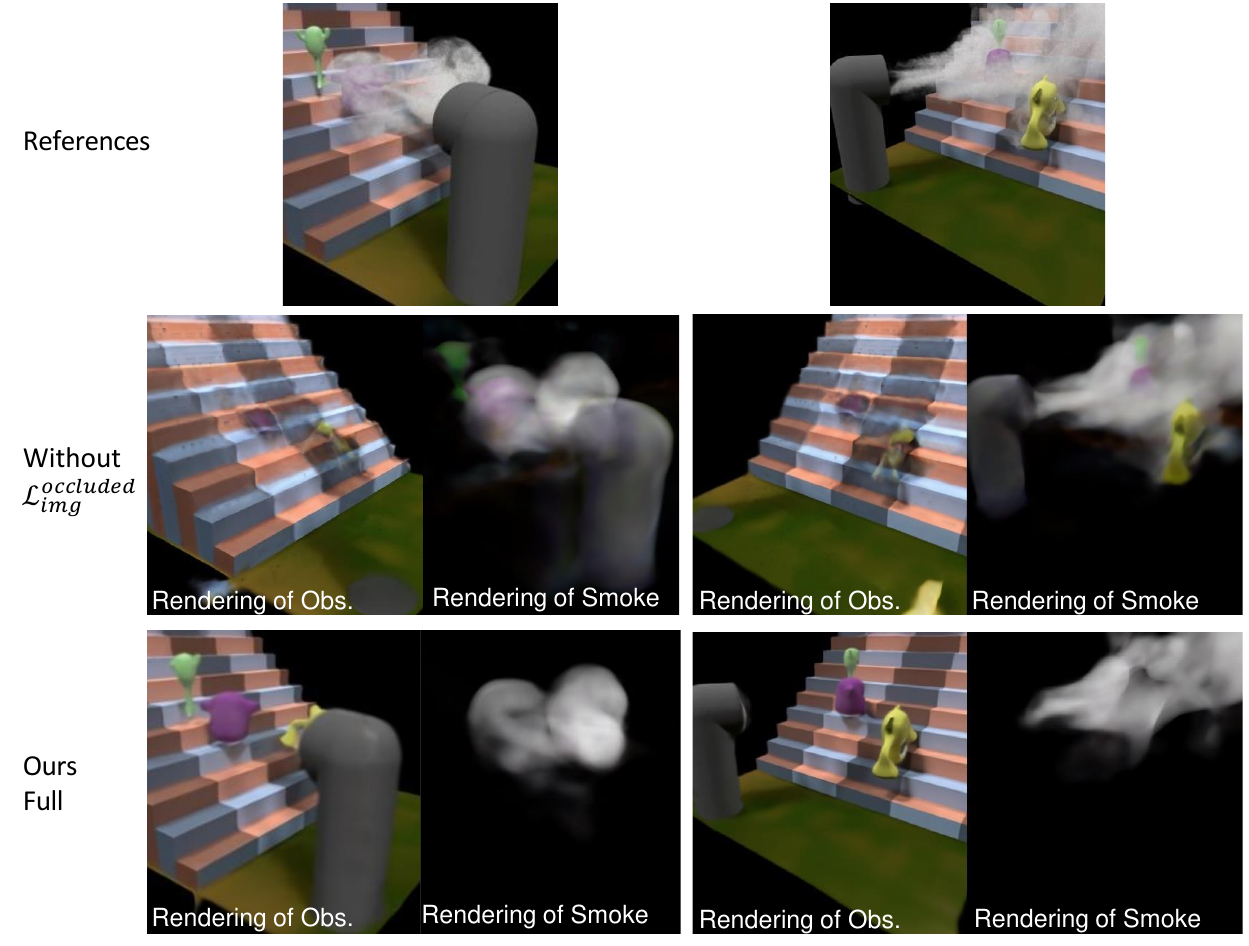}
    \vspace{-12pt}
	\caption
	{
        Without the proposed self-supervised scene decomposition constraints, static obstacles are likely to be modeled by the dynamic NeRF, leading to decomposition artifacts.
	}
	\label{fig:illustration_reconstruction}
\end{figure}
%
%

%% file: fig/comparison/fig_decomposed_novel_view_comp.tex
\begin{figure*}[t]
	\includegraphics[width=1.0\linewidth]{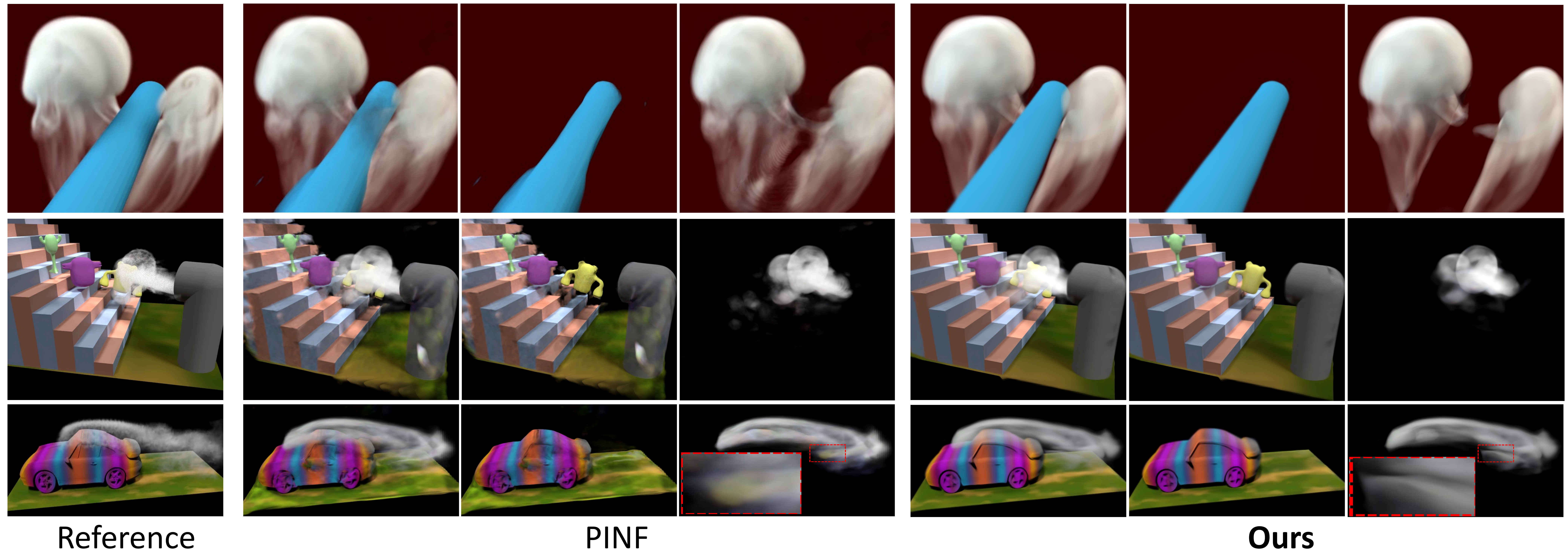}
    \vspace{-18pt}
	\caption
	{
     Comparison of the decomposed reconstruction of hybrid scenes. We compare our method with the state-of-the-art method PINF~\cite{chu2022pinf} supporting hybrid scene reconstruction with obstacles. Novel view synthesis results are rendered for the full scene, static obstacles only, and dynamic smoke only, demonstrating that our method achieves significantly better scene decomposition and more detailed reconstruction.
	}
	\label{fig:comparison_decomposed_novel_view_comp}
\end{figure*}

%% file: tex/Sec6_Experiments.tex
\section{Experiments}

\begin{table}[t]
\small
\renewcommand\arraystretch{0.5}
\caption{
Novel view comparisons of hybrid scenes. 
}
\label{table:novel_view_metrics}
\vspace{-12pt}
\begin{center}
\begin{tabular}{lcccccc}
\toprule
     &
  \multicolumn{3}{c}{PINF}&
 \multicolumn{3}{c}{\textbf{Ours}}\\
\midrule
 Dataset & 
 PSNR$\uparrow$&  SSIM$\uparrow$ & LPIPS$\downarrow$ & 
PSNR$\uparrow$&  SSIM$\uparrow$ & LPIPS$\downarrow$ \\
\midrule
Cylinder & 25.01 &  0.9097 & 0.1611  & \textbf{28.28} & \textbf{0.9318} &\textbf{0.1473} \\     
Game & 21.72 &   0.8175 & 0.3187   &\textbf{26.99} &  \textbf{0.8838} &\textbf{0.2140} \\ 
Car & 22.89  & 0.8731  & 0.1778   & \textbf{24.45} & \textbf{0.9124} &\textbf{0.1481} \\   
\midrule
Mean & 23.21 & 0.8668 & 0.2192 & \textbf{26.57} & \textbf{0.9093} & \textbf{0.1698}
 \\
\bottomrule
\end{tabular}
\end{center}
\end{table}

We first show results on three synthetic hybrid scenes with obstacles and complex light settings in Sec~\ref{sec:exp_hybrid_scenes}, then show results on real fluid captures in Sec~\ref{sec:exp_real_captured}. 
Ablation studies (Supp.~\ref{sec:exp_ablation}) and experimental details (Supp.~\ref{sec:exp_setting}) are presented in the supplement.

\subsection{Synthetic Scenes with Obstacles} \label{sec:exp_hybrid_scenes}
PINF~\cite{chu2022pinf} was the only method capable of reconstructing hybrid scenes with obstacles before us.
In the following, we compare ours with 
PINF quantitatively and qualitatively. 
We present a basic test case, called the Cylinder scene, and select two complex scenes in PINF, namely the Game scene and the Car scene.

\vspace{2pt}\noindent
\paragraph{Appearance and geometry reconstruction}
To illustrate the proficiency of our method in effectively reconstructing decomposed static obstacles and dynamic fluids from sparsely posed RGB videos of hybrid scenes, we first show a qualitative comparison of novel view synthesis results between our method and PINF in Fig.~\ref{fig:comparison_decomposed_novel_view_comp}. The results demonstrate that our method outperforms significantly in both the decomposition quality and the reconstructed details. 

Fig.~\ref{fig:comparison_mesh_recon} presents a qualitative comparison of the reconstructed obstacle geometry. Benefiting from our SDF representation of obstacles, our method can reconstruct an intact and detailed surface of the obstacle, an essential guarantee for plausible boundary constraints in dynamics reconstruction. On the contrary, it was infeasible to apply boundary constraints based on PINF due to the noisy obstacle surface extracted from its volumetric representation.
For quantitative evaluation, we measure the signal-to-noise ratio (PSNR), the structural similarity index measure (SSIM)~\cite{SSIM}, and the perceptual metric LPIPS~\cite{zhang2018lpips} between references and synthesized images of novel views, as shown in Table~\ref{table:novel_view_metrics}. The results indicate that our method significantly outperforms PINF in all measurements across three scenes.
\input{tab/tab_merge}

\input{fig/comparison/fig_vel_comp_cyl}
\input{fig/comparison/fig_merge_car_vel_and_trajectory}

\vspace{2pt}\noindent
\paragraph{Physical attributes reconstruction}
    To show a qualitative comparison between our method and PINF in terms of physical attributes reconstruction, we visualize the density, velocity, and vorticity of the reconstructed fluid from the front, side, and top views, as shown in Fig.~\ref{fig:comparison_vel_cyl},~\ref{fig:comparison_vel_car}, and~\ref{fig:comparison_vel_game}. 
Specifically, density volume is rendered with uniform ambient lighting, while the fluid velocity and vorticity are visualized using the middle slices of three views.
PINF shows blurred reconstruction results lacking high-frequency details, which is especially obvious in its vorticity visualizations.

\input{fig/comparison/fig_merge_mesh_scalar_recon}
We perform a quantitative evaluation measuring the voxel-wise $l_2$ error against the ground truth, in comparison to PINF~\cite{chu2022pinf}.
As shown in Table~\ref{table:vel_metrics}, our method exhibits substantially lower error rates in terms of reconstructed density and velocity, along with smaller divergence.  
This indicates significant improvements over reconstructed physical attributes using our approach.

Our trajectory representation enables efficient pathline visualization of fluid motion.
Fig.~\ref{fig:comparison_trajectory} randomly samples particles and visualizes their pathlines by connecting historical and future positions calculated by trajectory network inference. Pathlines closely follow the temporal evolution of the smoke, demonstrating valid long-term correspondence that was previously challenging to establish.

\subsection{Real Captures of Plume Scenes} ~\label{sec:exp_real_captured}

Aiming to validate the applicability in practical scenarios, we conduct experimentation on real fluid scenes. 
We use the ScalarFlow dataset~\cite{eckert2019scalarflow} which contains real fluid captures of rising smoke plumes. They use five fixed cameras distributed evenly in the front to capture the centered rising smoke. We take the middle 120 frames from each camera view for our fluid reconstruction task. Since no ground truth is available, we show qualitative results.
We begin our comparison by showcasing both a training view and a novel view to contrast our reconstruction results with PINF and HyFluid~\cite{yu2023inferring}. As depicted in Fig.~\ref{fig:comparison_scalar_recon}, our method excels in reconstructing markedly more detailed and coherent results. We also calculate the Frechet Inception Distance (FID) score between the novel view synthesis images and all training images. Our results achieve the best FID score of 166.6, followed by PINF’s 172.8 and Hyfluid’s 187.7.
Fig.~\ref{fig:comparison_vel_scalar} presents the comparison of reconstructed physical attributes. Our method demonstrates a notable reduction in noise levels within the density reconstruction, yielding a clearer and more continuous density distribution. Furthermore, the velocity and vorticity reconstructions achieved through our approach exhibit enhanced detail and clarity.

We refer readers to our supplemental video, which more clearly displays the quality of the dynamic reconstruction, consisting of radiance and trajectory fields.
We also conduct ablation studies, detailed in the supplementary, to show the effects of constraints and modules introduced in our method, e.g., the impact of long-term mapping, the boundary constraints, as well as our dual-density NeRF for dynamic smoke representation.

%% file: tab/tab_merge.tex
\begin{table}[t]
\small
\renewcommand\arraystretch{0.5}
\begin{center}
\caption{
Physical attributes evaluated on synthetic scenes. The divergence of velocity should be as close to zero as possible.
}
\label{table:vel_metrics}
\vspace{-12pt}
\begin{tabular}{cccc}
\toprule
$l_2$ errors on &density$\downarrow$ 
& velocity$\downarrow$ 
& divergence$\downarrow$ \\
\midrule
PINF  & 0.0083 &  0.2781 & 0.0083  \\
Ours  & \textbf{0.0058} &  \textbf{0.0703} & \textbf{0.0034}  \\
\bottomrule
\end{tabular}
\end{center}
\end{table}

%% file: fig/comparison/fig_vel_comp_cyl.tex
%
%
\begin{figure*}[tp]\footnotesize
	\includegraphics[width=\linewidth]{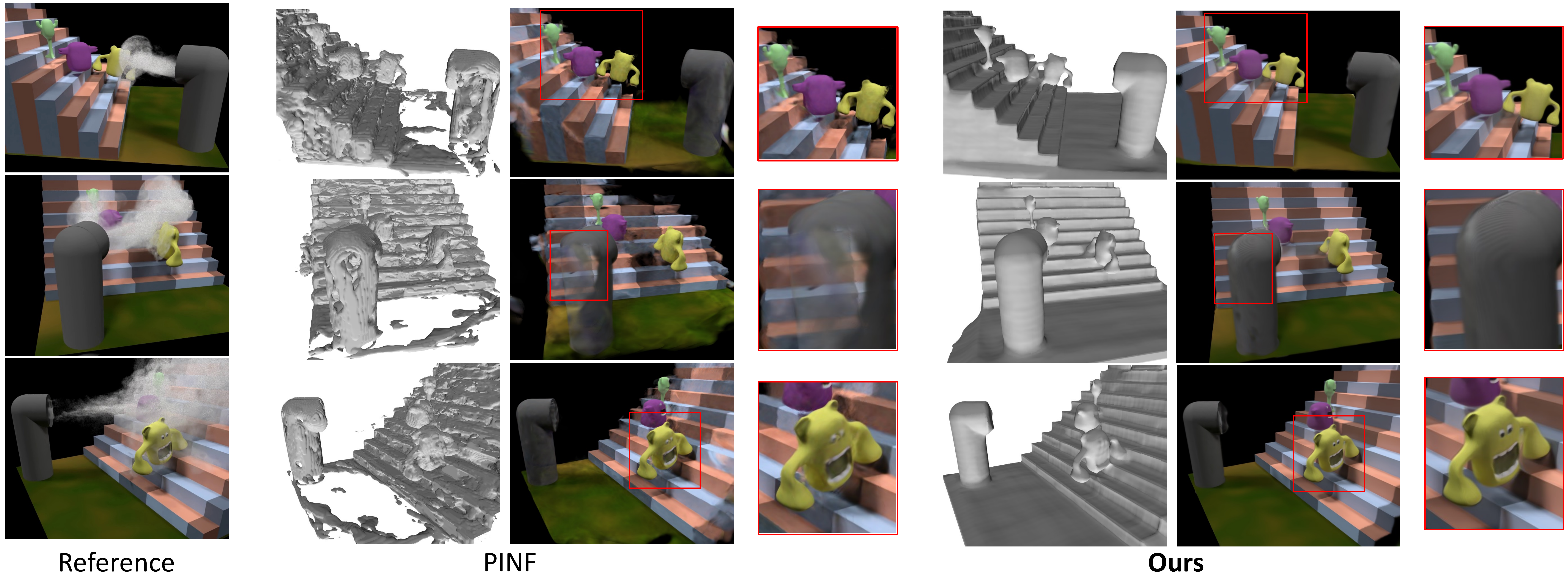}
	\caption
	{
     Comparison of the reconstruction of the scene's static component on the Game scene.
     We show both the geometry and novel view synthesis results. Our reconstruction results greatly outperforms PINF~\cite{chu2022pinf}, showing fine details without inducing noise. This high fidelity static obstacle reconstruction also lays a solid foundation for implementing accurate boundary constraints for dynamic fluid reconstruction.
	}
	\label{fig:comparison_mesh_recon}
     \vspace{6pt}
	\includegraphics[width=1.0\linewidth]{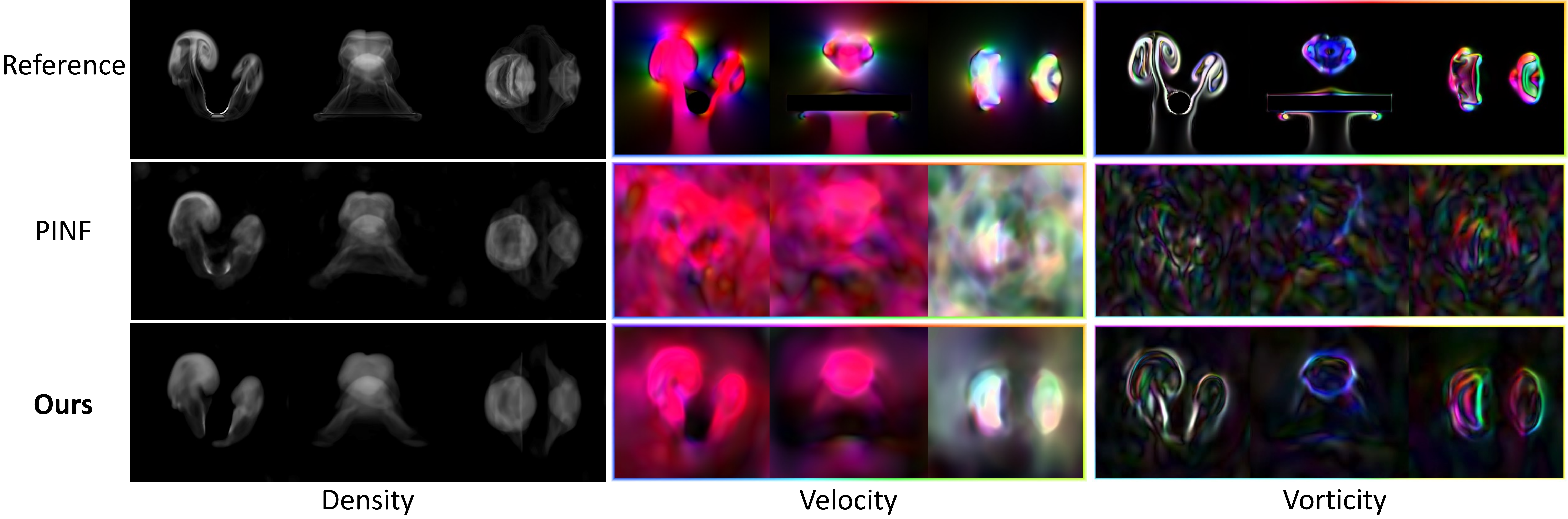}
	\caption
	{
		Comparison of the reconstruction of fluid's physical attributes on the Cylinder scene.
	}
	\label{fig:comparison_vel_cyl}
	%
\end{figure*}
%
%

%% file: fig/comparison/fig_merge_car_vel_and_trajectory.tex
\begin{figure*}[tp]\footnotesize
\begin{center}
    \begin{minipage}{.33\textwidth}
\centering
  	\includegraphics[width=\linewidth]{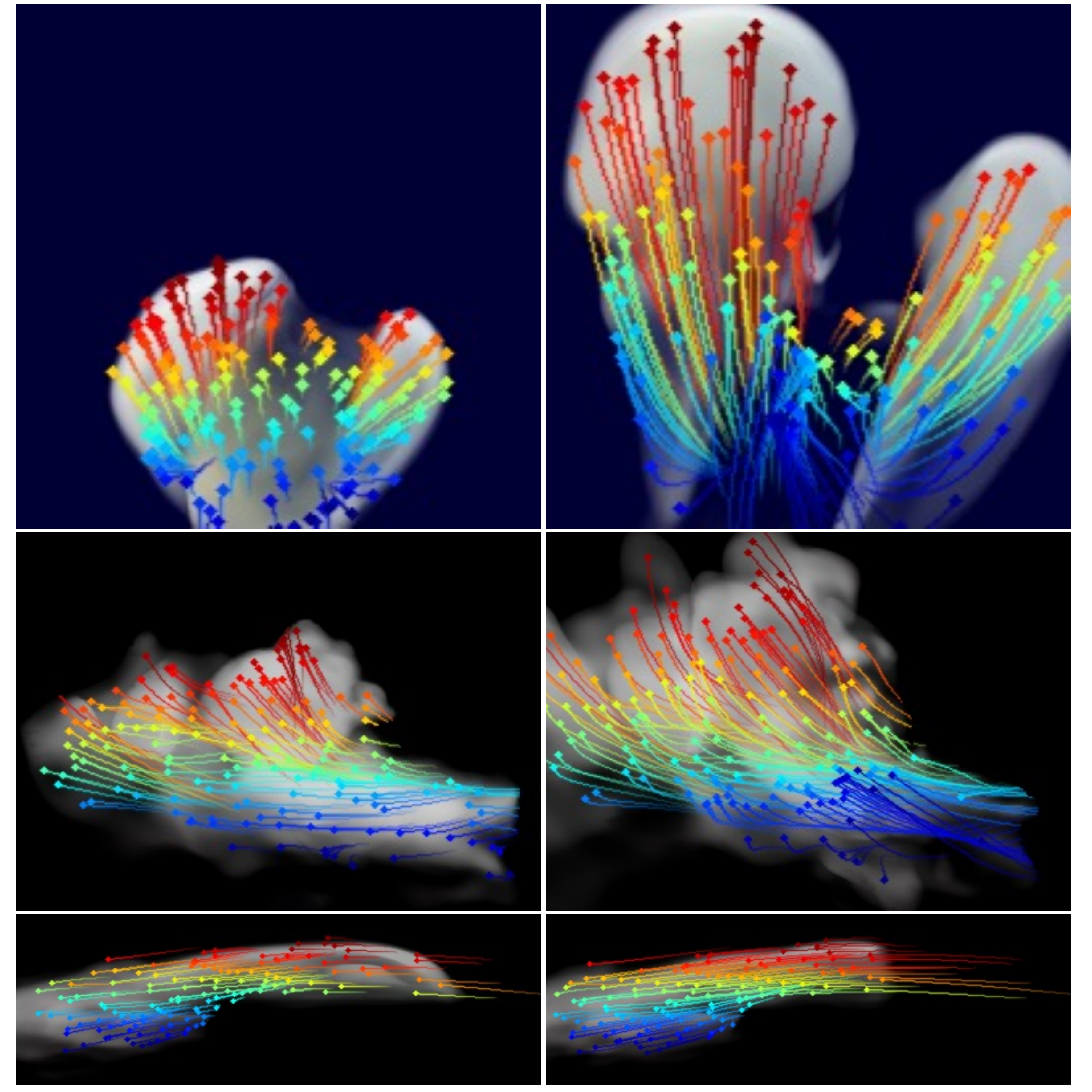}
	\caption
	{
        Trajectory visualization. 
        Please refer to the supplemental video for dynamic visualization.
	}
	\label{fig:comparison_trajectory}
 \end{minipage}
\hspace{8pt}
 \begin{minipage}{.62\textwidth}
 \centering
	\includegraphics[width=\linewidth]{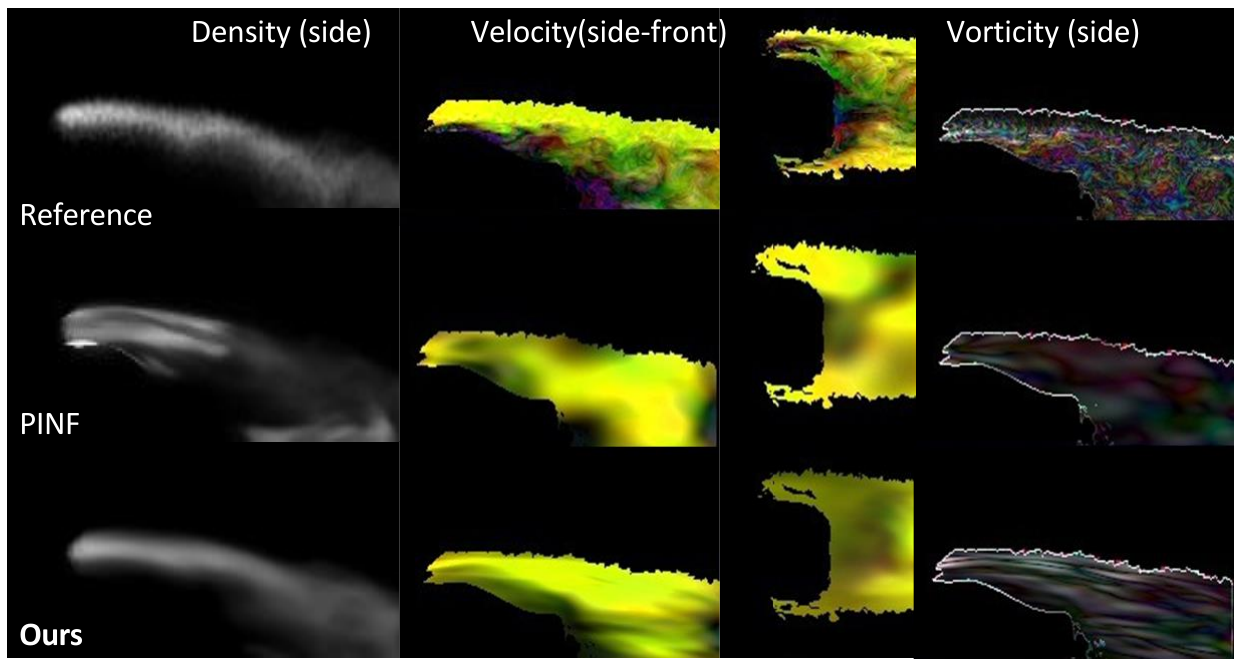}
	\caption
	{
		Comparison of physical attributes reconstructed on the Car scene.
		For better visualization, we show velocity and vorticity with a density mask for all methods.
        }
		\label{fig:comparison_vel_car}
 \end{minipage}

 \begin{minipage}{\textwidth}
\centering
 \vspace{18pt}
\includegraphics[width=\linewidth]{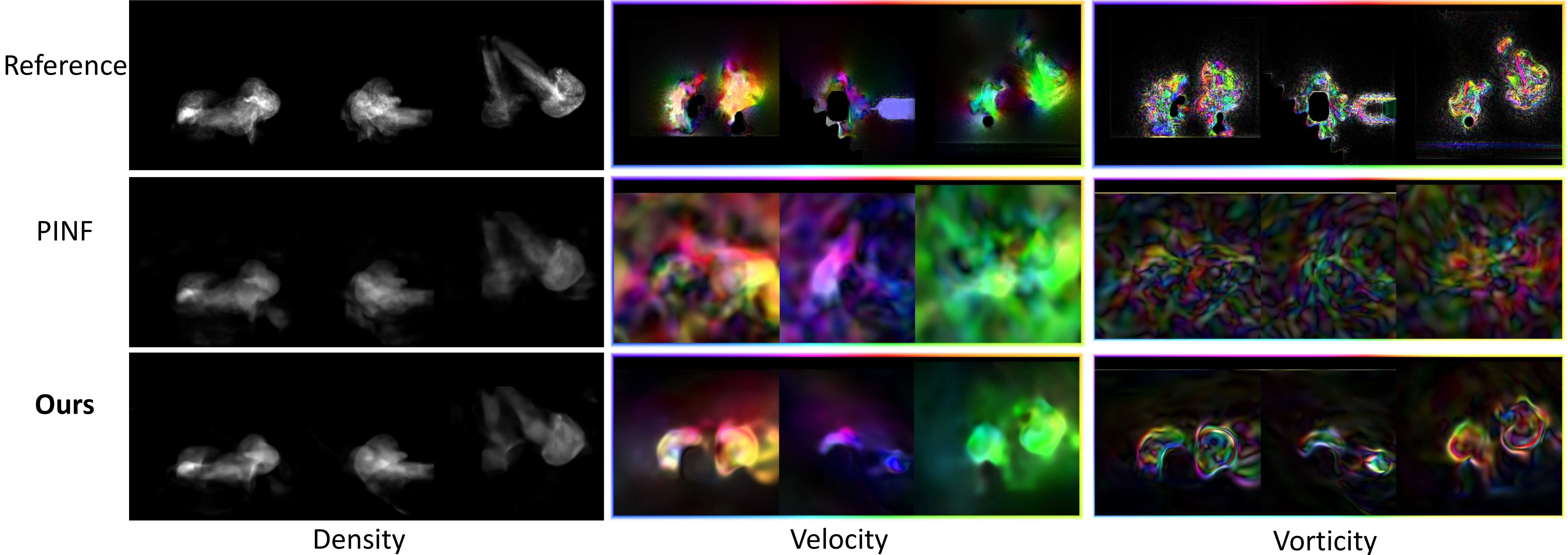}
	\caption
	{
		Reconstruction comparisons of the physical attributes on the Game scene.
	}
	\label{fig:comparison_vel_game}
\end{minipage}\\ 
\end{center}
\vspace{-6pt}
\clearpage
\end{figure*}

%% file: fig/comparison/fig_merge_mesh_scalar_recon.tex
\begin{figure*}[tp]\footnotesize
\begin{minipage}{.5\textwidth}
\centering
  	\includegraphics[width=\linewidth]{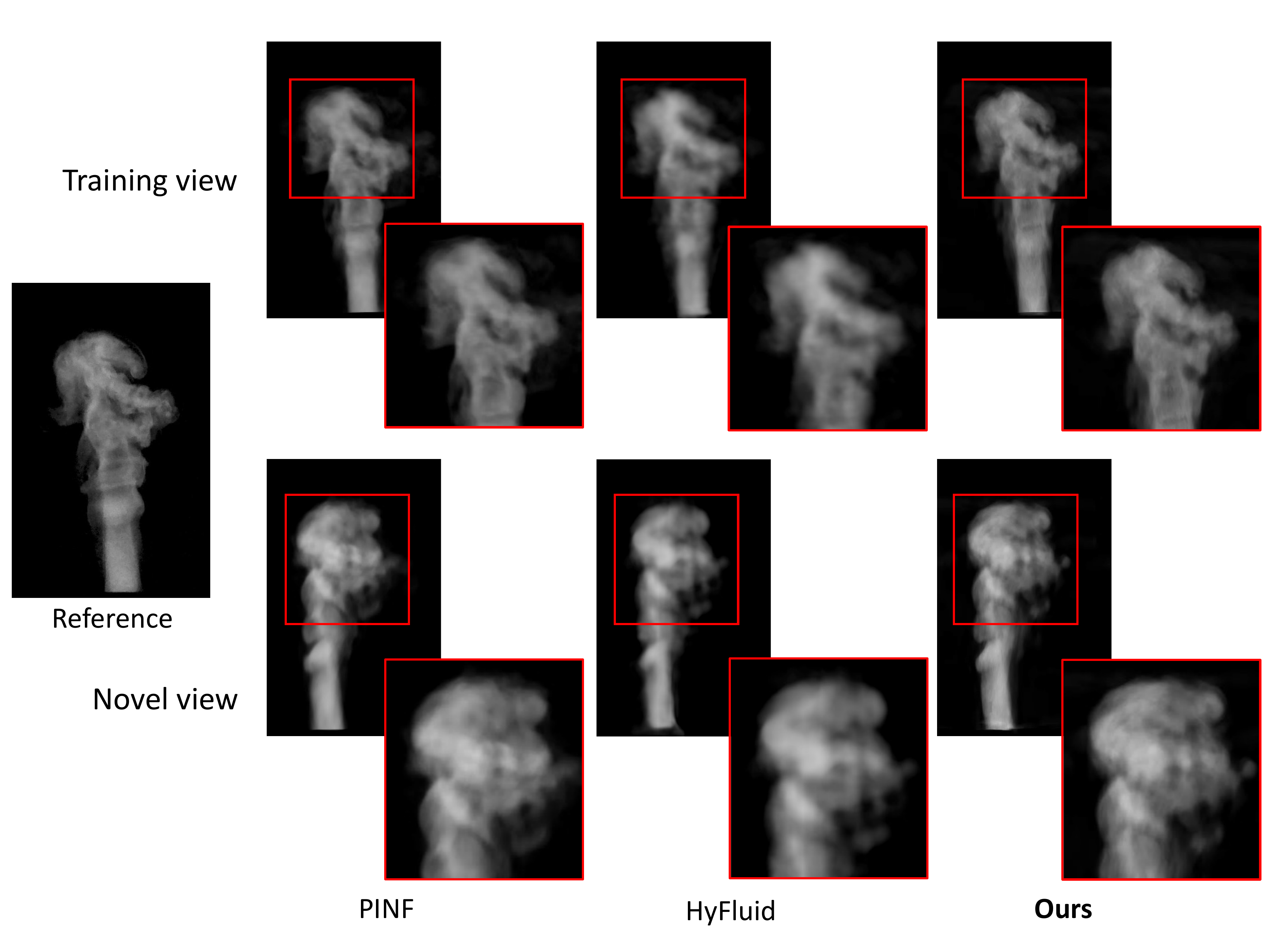}
   \vspace{-26pt}
	\caption
	{
     Comparison of the novel view synthesis results on a real capture. Our results show more details and are less overfitting.
	}
	\label{fig:comparison_scalar_recon}
 \end{minipage}\vspace{-2pt}
 \hfill
 \begin{minipage}{.47\textwidth}
 \centering
 	\includegraphics[width=0.95\linewidth]{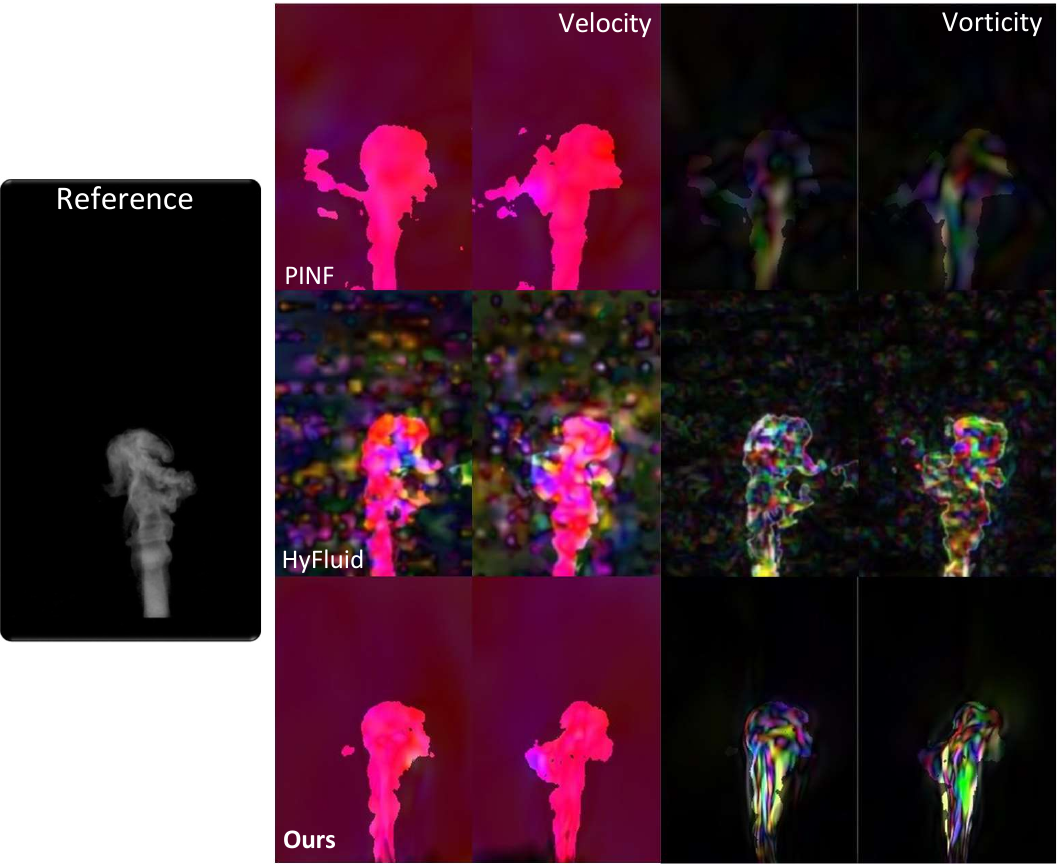}
    \vspace{-12pt}
	\caption
	{
		Comparison of the reconstruction of fluid's physical attributes on the real ScalarFlow captures. Our results contain less noise but sharper vorticity.
		}
	\label{fig:comparison_vel_scalar}
 \end{minipage}\vspace{-2pt}
\end{figure*}

%% file: tex/Sec7_Conclusion.tex
\section{Limitations}
Under the challenging sparse view setting, our method encounters limitations of insufficient ability to capture high-frequency details compared to references.
The main reason is that our work is based on NeRF and current NeRF methods face challenges in capturing high-frequency information under sparse viewpoint inputs. 
The situation becomes more challenging in tasks involving volumetric fluid, which forms the reconstruction bottleneck shared by our work as well as previous works like PINF~\cite{chu2022pinf} and Hyfluid~\cite{yu2023inferring}.
Another reason for the lack of details is that we have employed a relatively small model for our trajectory encoder and decoder due to the time complexity associated with auto-differentiation. Improving scalability and enhancing optimization efficiency is of interest. This could potentially be achieved by applying high-order differentiable network architectures~\cite{shen2022hod} and optimizing with more effective gradient descent involving multiple learning objectives~\cite{dong2022gdod}.
Applying 3D Gaussian splatting techniques~\cite{kerbl2023gaussian_splatting} and recent advancements in reconstruction under sparse view settings holds promise for improving density reconstruction.

Since we assume inviscid fluid and minimize the impact of pressure differences in our fluid modeling, our method may fail to accurately reconstruct complex smoke or liquid behaviors.
Future work also involves exploring more accurate physics for reconstruction, including inflow and pressure estimation. Lastly, building upon the flexibility and stability exhibited by our approach, we are interested in exploring multi-physics phenomena.

\section{Conclusion}
We introduced the Neural Characteristic Trajectory Field for NeRF-based smoke reconstruction. 
Transferring Eulerian coordinates to Lagrangian trajectories, this novel trajectory representation enables efficient flow map calculations and velocity extraction. 
Building on this, we propose physics-informed trajectory learning, integration into NeRF-based scene reconstruction, and self-supervised static-dynamic disentanglement.
Our approach facilitates comprehensive supervision, covering long-term conservation, short-term physics priors, and boundary conditions. Addressing challenges like occlusion uncertainty, density-color ambiguity, and static-dynamic entanglements, our method exhibits enhanced fidelity of reconstructed scenes, providing a valuable tool for physics-informed neural reconstruction in complex hybrid scenarios.

%% file: tex/supp.tex
\twocolumn[
\begin{center}
{\Huge \sffamily
    Physics-Informed Learning of Characteristic Trajectories\newline for Smoke Reconstruction - Supplementary	}\newline
        \vspace{-20pt}
		\subsection*{ 
        \begin{center}
		{\LARGE Yiming Wang,} ETH Zurich, Switzerland
          \end{center}
         \begin{center}
        {\LARGE Siyu Tang,} ETH Zurich, Switzerland
        \end{center}
        \begin{center}
        {\LARGE Mengyu Chu,} Peking University \& SKL of General AI, China
        \end{center}
		}
\end{center}
  \vspace{12pt}
]

\noindent
We first explain implementation details on the differentiable hybrid rendering and our network architectures. 
Then, we present experimental details including ablation studies, settings of evaluations, and training hyper-parameters.

\vspace{12pt}
\section{Implementation details}
\subsection{Hybrid Rendering of Static SDF and Dynamic NeRF} \label{sec:hybrid_render}
To apply the differentiable volume rendering of NeRF~\cite{mildenhall2020nerf}, we use the unbiased weight function derived for NeuS~\cite{wang-nips2021-neus} to convert the SDF to a volume density field. By accumulating the SDF-based densities and colors of the sample points, we can compute the color $\hat{C}$ of the ray with the same approximation scheme as used in NeRF
\begin{equation}
    \begin{aligned}
    \label{eqn:neus_vol_rendering}
        \hat{C}(\mathbf{r})=\sum_{i=1}^N T_i\alpha_i \mathbf{c}_i ~\text{~~ and ~~}
        T_i=\prod_{j=1}^{i-1}\left(1-\alpha_j\right)
    \end{aligned}
\end{equation}
where $\alpha(t_i)$ is a discrete density value defined by
\begin{equation}
    \begin{aligned}
        \label{eqn:neus_sdf2alpha}
    \alpha_i=\max \left(\frac{\phi\left(s_i\right)-\phi\left(s_{i+1}\right)}{\phi\left(s_i\right)}, 0\right)
    \end{aligned}
\end{equation}
Here, $\phi_s(x)=se^{-sx}/(1+e^{-sx})^2$ and $s$ is a learnable parameter.

We use a depth-aware rendering strategy to render an image for our hybrid scene. For each pixel, we first conduct ray marching to get sample points for the dynamic and static volumes separately, followed by a volume rendering of the merged sample points sorted based on their depth. Assume that we have $n_d$ sample points $\{\mathbf{\mathbf{p_d}(t_i)}=\mathbf{o}+t_i\mathbf{v}|i=0,1,\dots, n_d-1\}$ for dynamic scenes and $n_s$ sample points $\{\mathbf{\mathbf{p_s}(t_i)}=\mathbf{o}+t_i\mathbf{v}|i=0,1,\dots, n_s-1\}$ for static scenes along the camera ray, where $\mathbf{o}$ is the center of the camera and $\mathbf{v}$ is the view direction. 
These samples points $\mathbf{p_d(t_i)}$ and $\mathbf{p_s(t_i)}$ are then merged into samples points $\mathbf{p(t_i)}$ sorted based on the depth $t_i$ for composite volume rendering:
\begin{equation}
    \begin{aligned} \label{eqn:composite_rendering}
        \hat{C}_{composite}(\mathbf{r}) = \sum_{i=1}^{n_d + n_s} T_i\alpha_i \mathbf{c}_i ,
    \end{aligned}
\end{equation}
where $\alpha_i$ is calculated using Eq.~\ref{eqn:neus_sdf2alpha} for static scene samples, and for dynamic scene samples, it is computed as $1-\exp (-\sigma_i \delta_i)$ in Eq.~\ref{eqn:vol_rendering_discreted}.

Instead of adopting the two stage importance sampling strategy used in \cite{mildenhall-ECCV2021-nerf, chu2022pinf}, we use the ray marching acceleration strategy used in \cite{muller-SIG2022-instantngp, wang-ICCV2023-neus2} and extend it to the dynamic scenes for efficient and accurate smoke reconstruction.
Specifically, we maintain an occupancy grid that roughly marks each voxel cell as empty or non-empty for the static scene and each frame of the dynamic scenes.
The occupancy grid can effectively guide the marching process by preventing sampling in empty spaces and, thus, accelerates the volume rendering process.
We periodically update the occupancy grids based on the SDF value and the density value predicted by the static and dynamic models. 

\subsection{Network Architectures}
We use the MLP with periodic activation functions proposed in SIREN~\cite{sitzmann2020siren} to model all the networks. For the proposed Neural Characteristic Trajectory Field, the trajectory encoder has 4 hidden layers with a hidden size of 128, and the trajectory decoder has 3 hidden layers with a hidden size of 128. The trajectory characteristic dimension is set to 16. Regarding the integration with Neural Radiance Field, we adopt the time-varying NeRF model used in PINF~\cite{chu2022pinf}. 
The Lagrangian density is obtained by employing the encoded trajectory characteristic as input, processed through a SIREN MLP with two hidden layers.

\section{Experiment Details}

\subsection{Ablation Studies} \label{sec:exp_ablation}

We conduct sufficient experiments to showcase the effects of different modules in our reconstruction pipeline.

Firstly, we demonstrate that by utilizing the mapping obtained from our Neural Lagrangian Trajectory Field for long-term conservation constraints
, the reconstructed fluid quality can be significantly enhanced. 
As shown in Fig.~\ref{fig:ablation_color_mapping}, the color mapping loss $L^{color}_{mapping}$ is essential for decomposed reconstruction, especially for correcting the appearance of fluid. Without the color mapping loss, the smoke can easily learn the color of the car since we only have RGB supervision from composed images.
We also evaluate the impact of density mapping loss and velocity mapping loss on fluid density and velocity reconstruction qualitatively and quantitatively. The results in Fig.~\ref{fig:ablation_vel} and Table~\ref{table:ablation_vel} show that by utilizing the long-term conservation information of fluid dynamics, we can get better reconstruction results with more temporal consistency.
The effect of boundary constraints is also demonstrated in Fig.~\ref{fig:ablation_vel} and Table~\ref{table:ablation_vel}, illustrating the necessity of accurate boundary constraints for fluid dynamics.
%

\input{fig/ablation/fig_ablation_color_mapping}
\input{fig/ablation/fig_ablation_vel}
\input{tab/tab_ablation}
\input{fig/ablation/fig_ablation_two_layer_density}
\input{fig/ablation/fig_mapping_frame_error}
%
We propose a dual-density representation for enhanced fluid reconstruction. 
As illustrated in Fig.~\ref{fig:illustration_two_layer_density}, the coarse-level Lagrangian density exhibits structural density gradients, while the fine-level NeRF density captures more high-frequency details. 
The density gradients play a crucial role in the transport constraint, converting optical information into vital physical information for velocity estimation. 
To assess the impact of this dual-density representation, we conduct an ablation study using either density alone. 
As shown in Fig.~\ref{fig:ablation_two_layer_density}, relying solely on the fine-level density results in suboptimal and blurry results similar to PINF. 
Employing only the coarse-level density with structural density gradients makes learning easier for the velocity network, but, as shown in the second row, leads to a lack of details in the reconstructed velocity. 
Thus, by combining this two-level density, our full model achieves robust reconstruction visualized in the last row.
We further investigate the efficacy of the proposed intrinsic constraints for training the Neural Characteristic Trajectory Field. 
Fig.~\ref{fig:ablation_intrinsic_constraints} computes the distance between the mapping position obtained through a single network inference of our Neural Characteristic Trajectory Field with the mapping position numerically integrating velocity values sampled from the derivatives of our trajectory field.
Theoretically, the two positions should be the same if the trajectory feature stays consistent along the trajectory and the integrating interval is small enough. 
As shown in Fig.~\ref{fig:ablation_intrinsic_constraints}, intrinsic constraints drive the network inference and the numerical integration closer, 
indicating a more accurate trajectory inference with more consistent features. This consistency is essential for incorporating other long-term mapping constraints.

\subsection{Testing Scene Settings} \label{sec:exp_setting}
In line with PINF~\cite{chu2022pinf}, we use \textit{Mantaflow}~\cite{mantaflow2018} to simulate dynamic fluids with obstacles, and use \textit{Blender}~\cite{blender}  to render the synthetic scenes.
The Cylinder scene uses 5 evenly posed cameras and a complex lighting combination including point lights, directional lights, and an environment map. 
It is simulated with a spatial resolution of $256^3$ and a time step of 1.0.
In Blender, its spatial size is $4m\times4m\times4m$ and we render 120 frames with 1 frame per simulation step.
While the Cylinder scene is a basic scene with a regular obstacle,
the Game scene and the Car scene from PINF are more complex, containing enhanced vorticity using the wavelet turbulence method~\cite{kim2008wavelet} and obstacles in complex shapes.
Their simulation resolutions are $512\times432\times408$ and $768\times192\times307$, respectively. The lighting and camera settings are similar to the Cylinder scene.

\subsection{Details of Training} \label{sec:training_detail}
The training is divided into two stages. In the first stage, we only train the time-varying NeRF and the Lagrangian density. In the second stage, we train all components jointly. The total loss of our training takes the form of 
\begin{equation}
    \mathcal{L}_\text{total}=\sum_i \mathcal{L}_i · w_i, 
\end{equation}
where each loss term $(\mathcal{L}_i)$ and their corresponding weights $(w_i)$ are summarized in Table~\ref{tab:loss_functions}.
We train our network using the Adam~\cite{kingma2014adam} optimizer with a learning rate of $5\times10^{-4}$ and we exponentially decrease the learning rate by 0.1 every 250k training iterations. The first stage lasts for 50k iterations and the second stage lasts for 350k. The training takes around 12 hours on a single A100 GPU.

Results for the baseline methods PINF~\cite{chu2022pinf} and HyFluid~\cite{yu2023inferring} are generated using the officially released codes and configurations, presented with the authors' permission.
Training PINF for the same training iterations takes around 13 hours on the same machine, and training HyFluid takes around 9 hours in total.

\begin{table}[htbp]\small
\centering
\caption{Summary of loss functions with respective weights.}
\label{tab:loss_functions}
\noindent\begin{tabular}{lcc}
\toprule
$\mathcal{L}_i$  & Eqn. & $w_i$ \\
\midrule
\multicolumn{3}{l}{---\textit{For Physics-Informed Learning~(Sec.\ref{sec:learn_NCTJ})}:} \\[4pt]
$\mathcal{L}_{cycle}^{self}$      & \ref{Loss:self_cycle_loss}  & 1.0  \\
$\mathcal{L}_{cycle}^{cross}$     & \ref{Loss:cross_cycle_loss} & 1.0 \\
$\mathcal{L}_{feature}$        & \ref{Loss::trajectory_feature_material_derivative} & 0.01 \\
$\mathcal{L}_{NSE}$           & \ref{Loss:physical_constraints} & 0.001 \\
$\mathcal{L}_{div}    $        & \ref{Loss:physical_constraints} & 0.001 \\[2pt]
\multirow{2}{*}{$\mathcal{L}_{transport}^{full}$} & \multirow{2}{*}{\ref{Loss:full_density_transport}} & $w_i$: 0.01\textasciitilde0.1\\
   &  & $\lambda$ : 0.1\textasciitilde 1.0 \\[2pt]

$\mathcal{L}_{mapping}^{velocity}$    & \ref{Loss:velocity_mapping_loss} & 0.01\textasciitilde0.1\\
$\mathcal{L}_{mapping}^{density}$      & \ref{Loss:density_mapping_loss} & 0.01\textasciitilde0.1\\
\multirow{2}{*}{$\mathcal{L}_{mapping}^{color}$} & \multirow{2}{*}{\ref{Loss:density_mapping_loss}} &
{\footnotesize 10.0 for the car scene with extremely sever color ambiguity}
\\
& & {\footnotesize 0.01 for all other scenes}\\
$\mathcal{L}_{bnd}^{\mathcal{U}}$   & \ref{eqn:vel_boundary_loss}  & 0.5 \\[8pt]
\multicolumn{3}{l}{---\textit{For Scenes with Obstacles~(Sec.\ref{sec:boundary_recon})}:}
\\[4pt]
$\mathcal{L}_{bnd}^{\sigma}$  & \ref{eqn:vel_boundary_loss} &  0.5 \\
$\mathcal{L}_{img}^{combined}$ & \ref{Loss:final_combined_rgb_loss} & $w_i=1.0$ and $\alpha: 1\xrightarrow[]{2000 \  iterations}0$   \\
\bottomrule
\end{tabular}
\end{table}

%% file: fig/ablation/fig_ablation_color_mapping.tex
%
%
\begin{figure}[tb]
	\includegraphics[width=\linewidth]{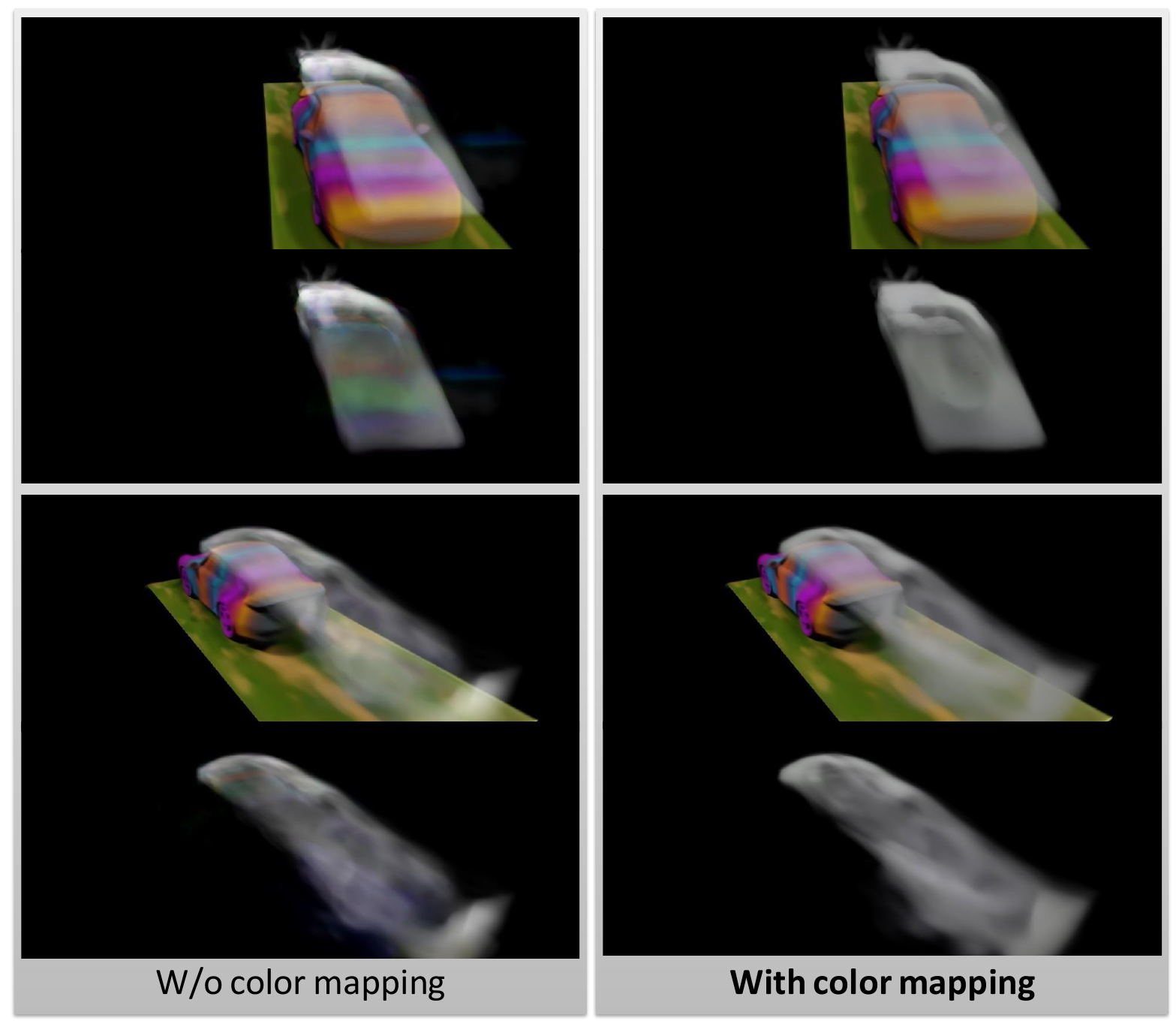}   
    \vspace{-24pt}
	\caption
	{
		Ablation study of the color mapping constraint. The color mapping constraint proves essential for decomposed reconstruction. Leveraging dynamics information, it prevents the density-color entanglement, ensuring that color of static obstacles does not bleed into the smoke.
	}
	\label{fig:ablation_color_mapping}
\end{figure}
%
%

%% file: fig/ablation/fig_ablation_vel.tex
%
%
\begin{figure}[tb]
	\includegraphics[width=\linewidth]{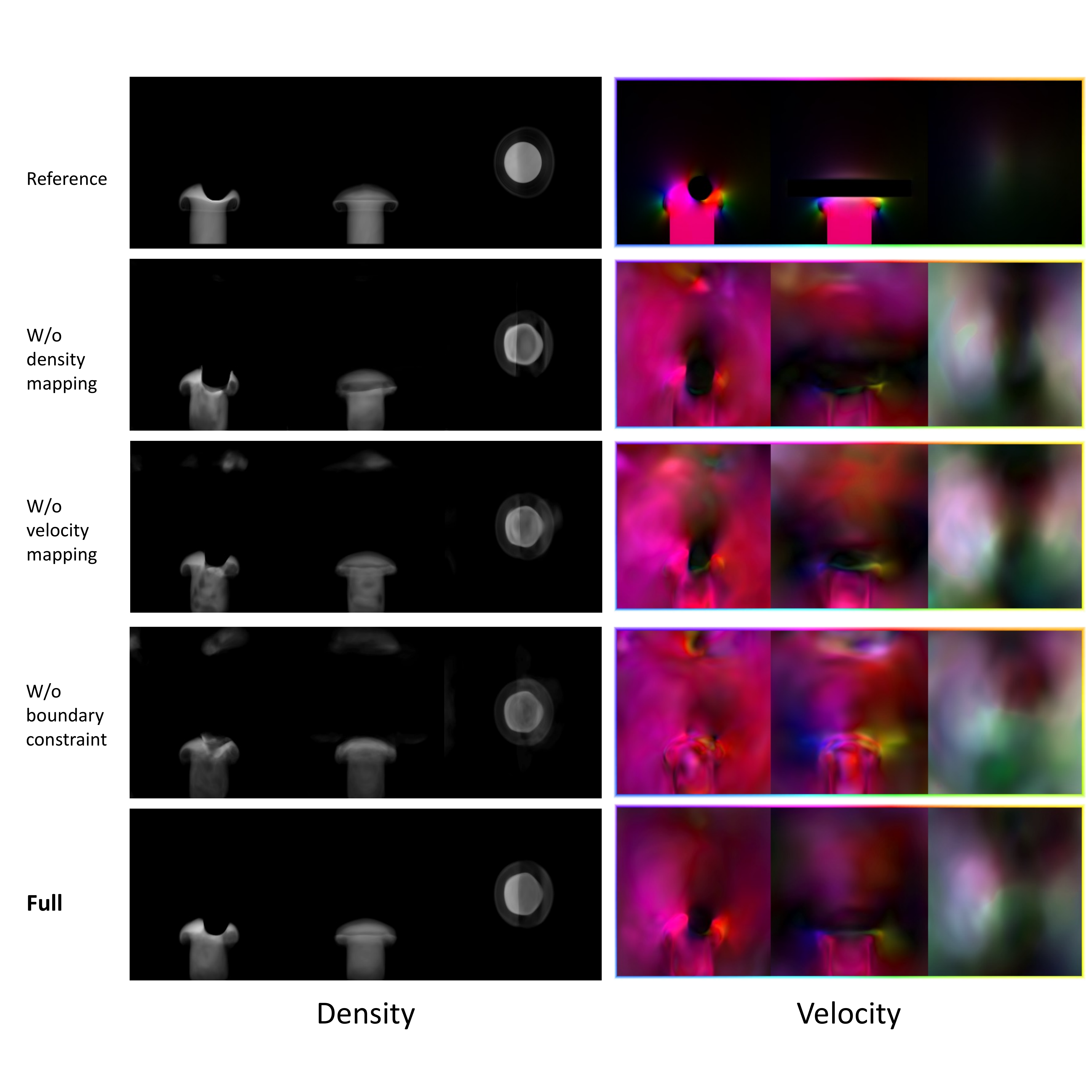}   
    \vspace{-36pt}
	\caption
	{
		Qualitative results of ablation study in using density mapping constraints, velocity mapping constraints, and boundary constraints.
	}
	\label{fig:ablation_vel}
	%
\end{figure}
%
%

%% file: tab/tab_ablation.tex
\begin{table}[tb]
\caption{
Quantitative results of ablation study in using density mapping constraints, velocity mapping constraints, and boundary constraints.
}
\label{table:ablation_vel}
\vspace{-12pt}
\begin{center}
\small
\begin{tabular}{lcc}
\toprule
 &density$\downarrow$ 
& velocity$\downarrow$  \\
\midrule
Ours (Full)  & \textbf{0.0058} &  \textbf{0.0703} \\
Ours (W/o density mapping)  & 0.0070 &  0.1287 \\
Ours (W/o velocity mapping)  & 0.0062 &  0.1469  \\
Ours (W/o boundary constraint) & 0.0072 &  0.1597  \\
\bottomrule
\end{tabular}
\end{center}
\end{table}

%% file: fig/ablation/fig_ablation_two_layer_density.tex
%
%
\begin{figure}
	%
	\includegraphics[width=\linewidth]{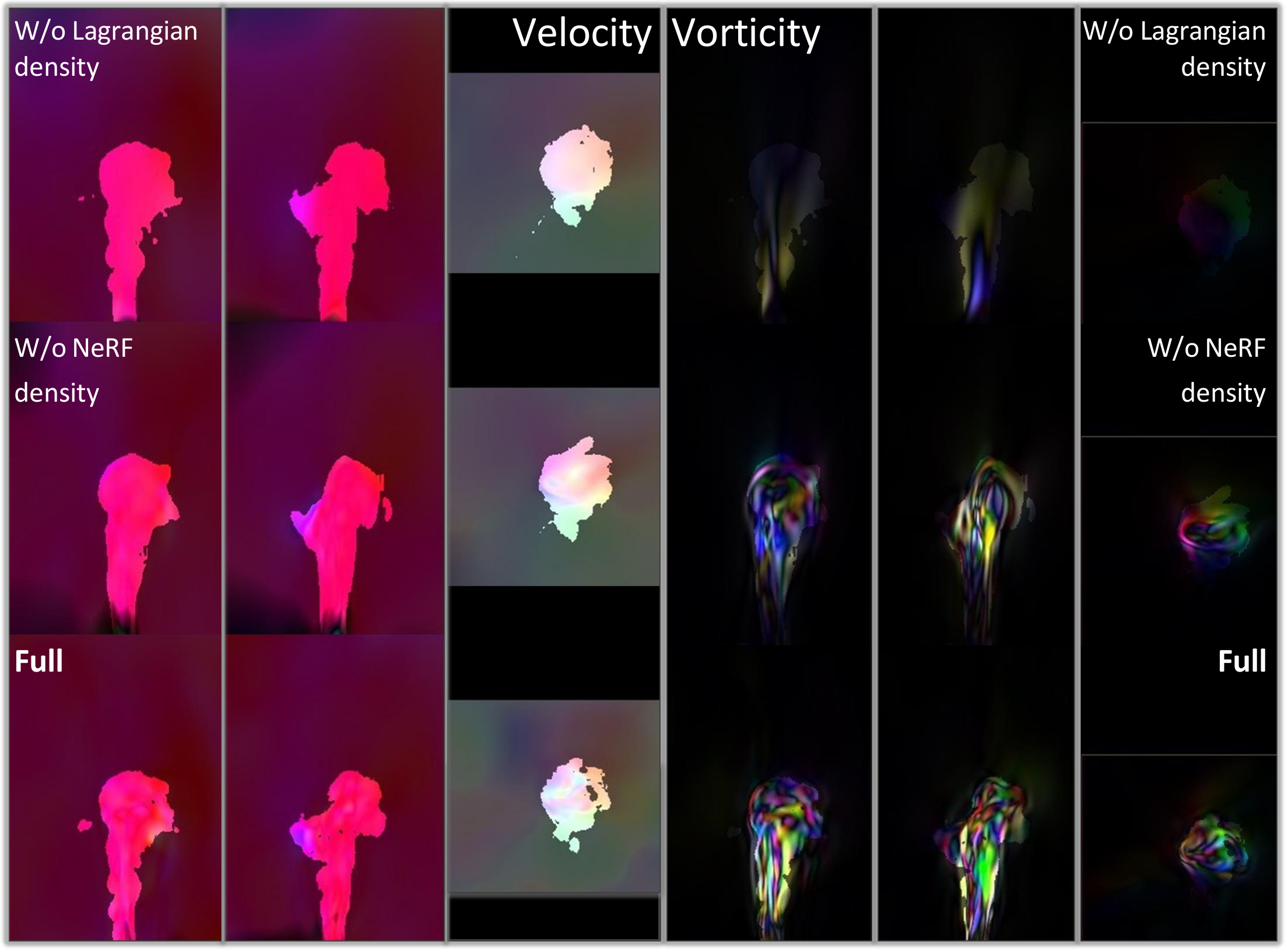}   
    \vspace{-18pt}
	\caption
	{
		Ablation study of using our dual density representation. Using NeRF density (fine level) along makes the velocity network hards to train, leading
		suboptimal and blurry results, while using Lagrangian density (coarse level) along lacks details in reconstructed velocity and vorticity.
	}
	\label{fig:ablation_two_layer_density}
	%
\end{figure}
%
%

%% file: fig/ablation/fig_mapping_frame_error.tex
%
%
\begin{figure}\footnotesize
	%
	\includegraphics[width=\linewidth]{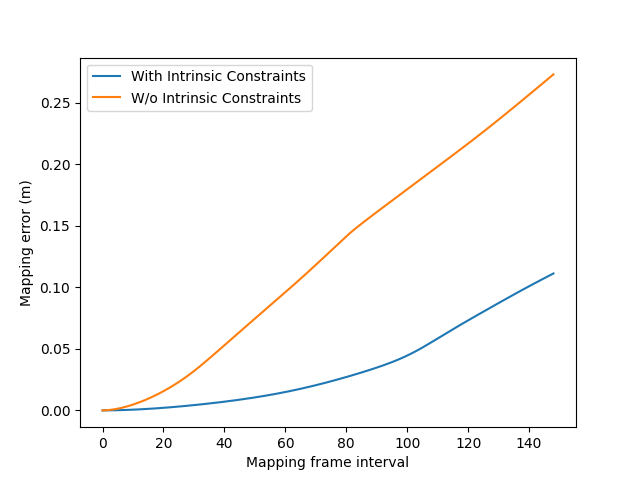}   
    \vspace{-18pt}
	\caption
	{
	    Ablation study for the intrinsic constraints for training Neural Lagrangian Trajectory Fields on the Cylinder scene. 
      Across different frame intervals, we calculate the distance between the mapping positions using single network inferences and the positions numerically integrated from velocity values.
      The Cylinder scene is in size of $4m \times 4m \times 4m$.
      The results demonstrate that our intrinsic constraints ensure accurate mapping, allowing long-term mapping constraints up to 50 steps in optimization.
	}
	\label{fig:ablation_intrinsic_constraints}
	%
\end{figure}
%
%